\documentclass[review]{elsarticle}

\usepackage{lineno,hyperref}
\modulolinenumbers[5]

\usepackage{graphicx}
\usepackage{subfigure}
\usepackage{url}
\usepackage{latexsym}
\usepackage{bm}
\usepackage{amsmath}
\usepackage{amssymb}
\usepackage{amsthm}
\usepackage{amsfonts}
\usepackage{algorithm}
\usepackage{algorithmic}
\usepackage{epstopdf}
\usepackage{float}
\usepackage{caption}
\usepackage{multirow}
\journal{an AI journal}

\begin{document}

\begin{frontmatter}

\title{Trainable back-propagated functional transfer matrices}


\author[address1]{Cheng-Hao Cai}
\ead{chenghao.cai@outlook.com}

\author[address2]{Yanyan Xu\corref{mycorrespondingauthor}}
\cortext[mycorrespondingauthor]{Corresponding author}
\ead{xuyanyan@bjfu.edu.cn}

\author[address3]{Dengfeng Ke}
\ead{dengfeng.ke@nlpr.ia.ac.cn}

\author[address4]{Kaile Su}
\ead{k.su@griffith.edu.au}

\author[address1]{Jing Sun}
\ead{j.sun@cs.auckland.ac.nz}

\address[address1]{Department of Computer Science, The University of Auckland, 38 Princes Street, Auckland 1142, New Zealand}
\address[address2]{School of Information Science and Technology, Beijing Forestry University, 35 Qing-Hua East Road, Beijing 100083, China}
\address[address3]{National Laboratory of Pattern Recognition, Institute of Automation, Chinese Academy of Sciences, 95 Zhong-Guan-Cun East Road, Beijing 100190, China}
\address[address4]{Institute for Integrated and Intelligent Systems, Griffith University, 170 Kessels Road, Nathan, QLD 4111, Australia}

\begin{abstract}
Connections between nodes of fully connected neural networks are usually represented by weight matrices. In this article, functional transfer matrices are introduced as alternatives to the weight matrices: Instead of using real weights, a functional transfer matrix uses real functions with trainable parameters to represent connections between nodes. Multiple functional transfer matrices are then stacked together with bias vectors and activations to form deep functional transfer neural networks. These neural networks can be trained within the framework of back-propagation, based on a revision of the delta rules and the error transmission rule for functional connections. In experiments, it is demonstrated that the revised rules can be used to train a range of functional connections: 20 different functions are applied to neural networks with up to 10 hidden layers, and most of them gain high test accuracies on the MNIST database. It is also demonstrated that a functional transfer matrix with a memory function can roughly memorise a non-cyclical sequence of 400 digits.
\end{abstract}


\end{frontmatter}

\linenumbers

\section{Introduction}	
\label{sec:intro}

Many neural networks use weights to represent connections between nodes. For instance, a fully connected deep neural network usually has a weight matrix in each hidden layer, and the weight matrix, a bias vector and a nonlinear activation are used to map input signals to output signals \cite{DBLP:journals/taslp/MohamedDH12}. Much work has been done on combining neural networks with different activations, such as the logistic function \cite{DBLP:journals/jbi/DreiseitlO02}, the rectifier \cite{DBLP:journals/jmlr/GlorotBB11}, the maxout function \cite{DBLP:conf/nips/MontufarPCB14} and the long short-term memory blocks \cite{DBLP:journals/neco/HochreiterS97}. In this work, we study neural networks from another viewpoint: Instead of using different activations, we replace weights in the weight matrix with functional connections. In other words, the connections between nodes are represented by using real functions, but no longer real numbers. Specifically, this work focuses on:
\begin{itemize}
\item \textbf{Extending the back-propagation algorithm to the training of functional connections.} The extended back-propagation algorithm includes rules for computing deltas of parameters in functional connections and rules for transmitting error signals from a layer to its previous layer. This algorithm is adapted from the standard back-propagation algorithm for fully connected feedforward neural networks \cite{DBLP:journals/nn/Hecht-Nielsen88a}.
\item \textbf{Discussing the meanings of some functionally connected structures.} Different functional transfer matrices can construct different mathematical structures. Although it is not necessary for these structures to have meanings, some of them do have meanings in practice.
\item \textbf{Discussing some practical training methods for (deep) functional transfer neural networks.} Although the theory of back-propagation is applicable for these functional models, it is difficult to train them in practice. Also, the training becomes more difficult as the depth increases. In order to train them successfully, some tricks may be required.
\item \textbf{Demonstrating that these functional transfer neural networks can practically work.} We apply these models to the MNIST hand-written digit dataset \cite{DBLP:journals/spm/Deng12}, in order to demonstrate that different functional connects can work for classification tasks. Also, we try to make a model with a memory function memorise the circumference ratio $ \pi $, in order to demonstrate that functional connects can be used as a memory block.
\end{itemize}

The rest of this paper is organised as follows: Section \ref{sec:dnn} provides a brief review of the standard deep feedforward neural networks. Section \ref{sec:fnn} introduces the theory of functional transfer neural networks, including functional transfer matrices and a back-propagation algorithm for functional connections. Section \ref{sec:func_exam} provides some examples for explaining the meanings of functionally connected structures. Section \ref{sec:training} discusses training methods for practical use. Section \ref{sec:exp} provides experimental results. Section \ref{sec:relatedwork} introduces related work. Section \ref{sec:con} concludes this work and discusses future work.

\section{Background on deep feedforward neural networks}
\label{sec:dnn}

We assume that readers have been familiar with deep feedforward neural networks and only mention concepts and formulae relevant to our research \cite{DBLP:journals/nn/Hecht-Nielsen88a}: A feedforward neural network usually consists of an input layer, some hidden layers and an output layer. Two neighbouring layers are connected by a linear weight matrix $ \bm{W} $, a bias vector $ \bm{b} $ and an activation $ \varphi{u} $. Given an vector $ \bm{x} $, the neural network can map it to another vector $ \bm{y} $ via:
\begin{equation}
\bm{y} = \varphi(\bm{W} \cdot \bm{x} + \bm{b})~~.
\end{equation}

\section{A theory of functional transfer neural networks}
\label{sec:fnn}

\subsection{Functional transfer matrices}

Instead of using weights, a functional transfer matrix uses trainable functions to represent connections between nodes. Formally, it is defined as:
\begin{equation}
\label{eq_fun_mat}
\bm{F}=
\begin{pmatrix}
F_{1,1}(x) & F_{1,2}(x) & \cdots & F_{1,n}(x) \\
F_{2,1}(x) & F_{2,2}(x) & \cdots & F_{2,n}(x) \\
\vdots & \vdots & \ddots & \vdots \\
F_{m,1}(x) & F_{m,2}(x)  & \cdots & F_{m,n}(x) \\
\end{pmatrix}
~~,
\end{equation}
where each $ F_{i,j}(x) $ is a trainable function. We also define a transfer computation ``$ \odot $" for the functional transfer matrix and a column vector $ \bm{v}=(v_1,v_2,\cdots,v_n)^T $ such that:
\begin{equation}
\label{def_fmat_vec}
\begin{aligned}
\bm{F} \odot \bm{v} &=
\begin{pmatrix}
F_{1,1}(x) & F_{1,2}(x) & \cdots & F_{1,n}(x) \\
F_{2,1}(x) & F_{2,2}(x) & \cdots & F_{2,n}(x) \\
\vdots & \vdots & \ddots & \vdots \\
F_{m,1}(x) & F_{m,2}(x)  & \cdots & F_{m,n}(x) \\
\end{pmatrix}
\odot
\begin{pmatrix}
v_1 \\
v_2 \\
\vdots \\
v_n \\
\end{pmatrix}
\\
&=
\begin{pmatrix}
F_{1,1}(v_1) + F_{1,2}(v_2) + \cdots + F_{1,n}(v_n) \\
F_{2,1}(v_1) + F_{2,2}(v_2) + \cdots + F_{2,n}(v_n) \\
\vdots \\
F_{m,1}(v_1) + F_{m,2}(v_2)  + \cdots + F_{m,n}(v_n) \\
\end{pmatrix}
~~.
\end{aligned}
\end{equation}
Suppose that $ \bm{b}=(b_1,b_2,\cdots,b_m)^T $ is a bias vector, $ \varphi(u) $ is an activation, and $ \bm{x}=(x_1,x_2,\cdots,x_n)^T $ is an input vector. An output vector $ \bm{y}=(y_1,y_2,\cdots,y_m)^T $ can then be computed via
\begin{equation}
\label{eq_rel_inp_out}
\bm{y} = \varphi(\bm{F} \odot \bm{x} + \bm{b})~~.
\end{equation}
In other words, the $i$th element of the output vector is
\begin{equation}
\label{eq_inp_out}
y_i=\varphi(\sum\limits_{j=1}^{n}F_{i,j}(x_j)+b_i)~~.
\end{equation}

\subsection{Back-propagation}
\label{sec:backp}

Multiple functional transfer matrices can be stacked together with bias vectors and activations to form a multi-layer model, and the model can be trained via back-propagation \cite{DBLP:journals/nn/Hecht-Nielsen88a}: Suppose that a model has $ L $ hidden layers. For the $ l $th hidden layer, $ \bm{F^{(l)}} $ is a functional transfer matrix, its function $ F^{(l)}_{i,j}(x) $ is a real function with an independent variable $ x $ and $ r $ trainable parameters $ p^{(l)}_{i,j,k} (k=1,2,\cdots,r)$, $ \bm{b^{(l)}}=(b^{(l)}_1,b^{(l)}_2,\cdots,b^{(l)}_m)^T $ is a bias vector, $ \varphi^{(l)}(u) $ is an activation, and $ \bm{x^{(l)}}=(x^{(l)}_1,x^{(l)}_2,\cdots,x^{(l)}_n)^T $ is an input signal. An output signal $ \bm{y^{(l)}}=(y^{(l)}_1,y^{(l)}_2,\cdots,y^{(l)}_m)^T $ is computed via:
\begin{equation}
\label{eq_rel_inp_out_l}
\bm{y^{(l)}} = \varphi^{(l)}(\bm{F^{(l)}} \odot \bm{x^{(l)}} + \bm{b^{(l)}})~~.
\end{equation}
The output signal is the input signal of the next layer. In other words, if $ l \leq L $, then $ \bm{y^{(l)}} = \bm{x^{(l+1)}} $. After an error signal is evaluated by an output layer and an error function (see Section \ref{sec:train_single}), the parameters can be updated: Suppose that $ \bm{\delta^{(l)}_m} = (\delta^{(l)}_1,\delta^{(l)}_2,\cdots,\delta^{(l)}_m)^T $ is an error signal of output nodes, and $ \epsilon $ is a learning rate. Let $ u^{(l)}_i = \sum\limits_{j=1}^{n}F^{(l)}_{i,j}(x^{(l)}_j)+b^{(l)}_i $. Deltas of the parameters are computed via:
\begin{equation}
\label{eq_del_par}
\begin{aligned}
\Delta p^{(l)}_{i,j,k}&=\epsilon \cdot \delta^{(l)}_i \cdot \cfrac{\partial y^{(l)}_i}{\partial p^{(l)}_{i,j,k}} \\
&=\epsilon \cdot \delta^{(l)}_i \cdot
\cfrac{\partial y^{(l)}_i}{\partial u^{(l)}_i} \cdot \cfrac{\partial u^{(l)}_i}{\partial F^{(l)}_{i,j}(x^{(l)}_j)} \cdot \cfrac{\partial F^{(l)}_{i,j}(x^{(l)}_j)}{\partial p^{(l)}_{i,j,k}} \\
&=\epsilon \cdot \delta^{(l)}_i \cdot \cfrac{\partial \varphi^{(l)}(u^{(l)}_i)}{\partial u^{(l)}_i} \cdot \cfrac{\partial F^{(l)}_{i,j}(x^{(l)}_j)}{\partial p^{(l)}_{i,j,k}}~~.
\end{aligned}
\end{equation}
Deltas of the biases are the same as the conventional rule:
\begin{equation}
\label{eq_del_bia}
\begin{aligned}
\Delta b^{(l)}_i
&=\epsilon \cdot \delta^{(l)}_i \cdot \cfrac{\partial y^{(l)}_i}{\partial b^{(l)}_i} \\
&=\epsilon \cdot \delta^{(l)}_i \cdot \cfrac{\partial y^{(l)}_i}{\partial u^{(l)}_i} \cdot \cfrac{\partial u^{(l)}_i}{\partial b^{(l)}_i} \\
&= \epsilon \cdot \delta^{(l)}_i \cdot \cfrac{\partial \varphi^{(l)}(u^{(l)}_i)}{\partial u^{(l)}_i}~~.
\end{aligned}
\end{equation}
An error signal of the $(l-1)$th layer is computed via:
\begin{equation}
\label{eq_err_pre}
\begin{aligned}
\delta^{(l-1)}_j &= \sum\limits_{i=1}^{m} \delta^{(l)}_i \cdot \cfrac{\partial y^{(l)}_i}{\partial x^{(l)}_j} \\
&= \sum\limits_{i=1}^{m}  \delta^{(l)}_i \cdot \cfrac{\partial y^{(l)}_i}{\partial u^{(l)}_i} \cdot \cfrac{\partial u^{(l)}_i}{\partial F^{(l)}_{i,j}(x^{(l)}_j)} \cdot \cfrac{\partial F^{(l)}_{i,j}(x^{(l)}_j)}{\partial x^{(l)}_j} \\
& = \sum\limits_{i=1}^{m} \delta^{(l)}_i \cdot \cfrac{\partial \varphi^{(l)}(u^{(l)}_i)}{\partial u^{(l)}_i} \cdot \cfrac{\partial F^{(l)}_{i,j}(x^{(l)}_j)}{\partial x^{(l)}_j}~~.
\end{aligned}
\end{equation}
Please note that the computation of $ \cfrac{\partial F^{(l)}_{i,j}(x^{(l)}_j)}{\partial x^{(l)}_j} $
can be simplified as a transfer computation: Let $ \bm{F'^{(l)}} $ be a derivative matrix in which each element is $ \cfrac{\partial F^{(l)}_{i,j}(x)}{\partial x} $. Then the computation can be done via $ \bm{F'^{(l)}} \odot \bm{x^{(l)}} $. Similarly, the computation of $ \cfrac{\partial F^{(l)}_{i,j}(x^{(l)}_j)}{\partial p^{(l)}_{i,j,k}} $ in Eq. (\ref{eq_del_par}) can also be simplified: Let $ \bm{F'^{(l)}_{k}} $ be a derivative matrix in which each element is $ \cfrac{\partial F^{(l)}_{i,j}(x)}{\partial p_{i,j,k}} $. Then the computation can be done via $ \bm{F'^{(l)}_k} \odot \bm{x^{(l)}} $. Both $ \bm{F'^{(l)}} $ and $ \bm{F'^{(l)}_{k}} $ can be computed symbolically before applying the transfer computation.

\section{Examples}
\label{sec:func_exam}

\subsection{Periodic functions}
\label{sec:per_func}

Periodic functions can be applied to functional transfer matrices. For instance, $ f(x) = A \cdot sin(\omega \cdot x + \mu) $ is a periodic function, where $ A $, $ \omega $ and $ \mu $ are its amplitude, angular velocity and initial phase respectively. Thus, we can define a matrix consisting of the following function:
\begin{equation}
\label{sine_funcmat}
F_{i,j}(x) = p_{i,j,1} \cdot sin(p_{i,j,2} \cdot x + p_{i,j,3})~~,
\end{equation}
where $ p_{i,j,1} $, $ p_{i,j,2} $ and $ p_{i,j,3} $ are trainable parameters. If it is an $ M \times N $ matrix, then it can transfer an $ N $-dimensional vector $ (v_1,v_2,\cdots,v_N)^T $ to an $ M $-dimensional vector $ (u_1,u_2,\cdots,u_M)^T $ such that $ u_i =  \sum\limits_{j=1}^{N} p_{i,j,1} \cdot sin(p_{i,j,2} \cdot v_j + p_{i,j,3}) $. It is noticeable that each $ u_i $ is a composition of many periodic functions, and their amplitudes, angular velocities and initial phases can be updated via training. The training process is based on the following partial derivatives:
\begin{equation}
\cfrac{\partial F_{i,j}(x)}{\partial p_{i,j,1}} = sin(p_{i,j,2} \cdot x + p_{i,j,3})~~,
\end{equation}
\begin{equation}
\cfrac{\partial F_{i,j}(x)}{\partial p_{i,j,2}} = p_{i,j,1} \cdot cos(p_{i,j,2} \cdot x + p_{i,j,3}) \cdot x~~,
\end{equation}
\begin{equation}
\cfrac{\partial F_{i,j}(x)}{\partial p_{i,j,3}} = p_{i,j,1} \cdot cos(p_{i,j,2} \cdot x + p_{i,j,3})~~,
\end{equation}
\begin{equation}
\label{ell_fd}
\cfrac{\partial F_{i,j}(x)}{\partial x} = p_{i,j,1} \cdot p_{i,j,2} \cdot cos(p_{i,j,2} \cdot x + p_{i,j,3})~~,
\end{equation}
Similarly, this method can be extended to the cosine function.

\subsection{Modelling of conic hypersurfaces}
\label{sec:conic_func}

\begin{figure}
\centering
\includegraphics[width=7cm]{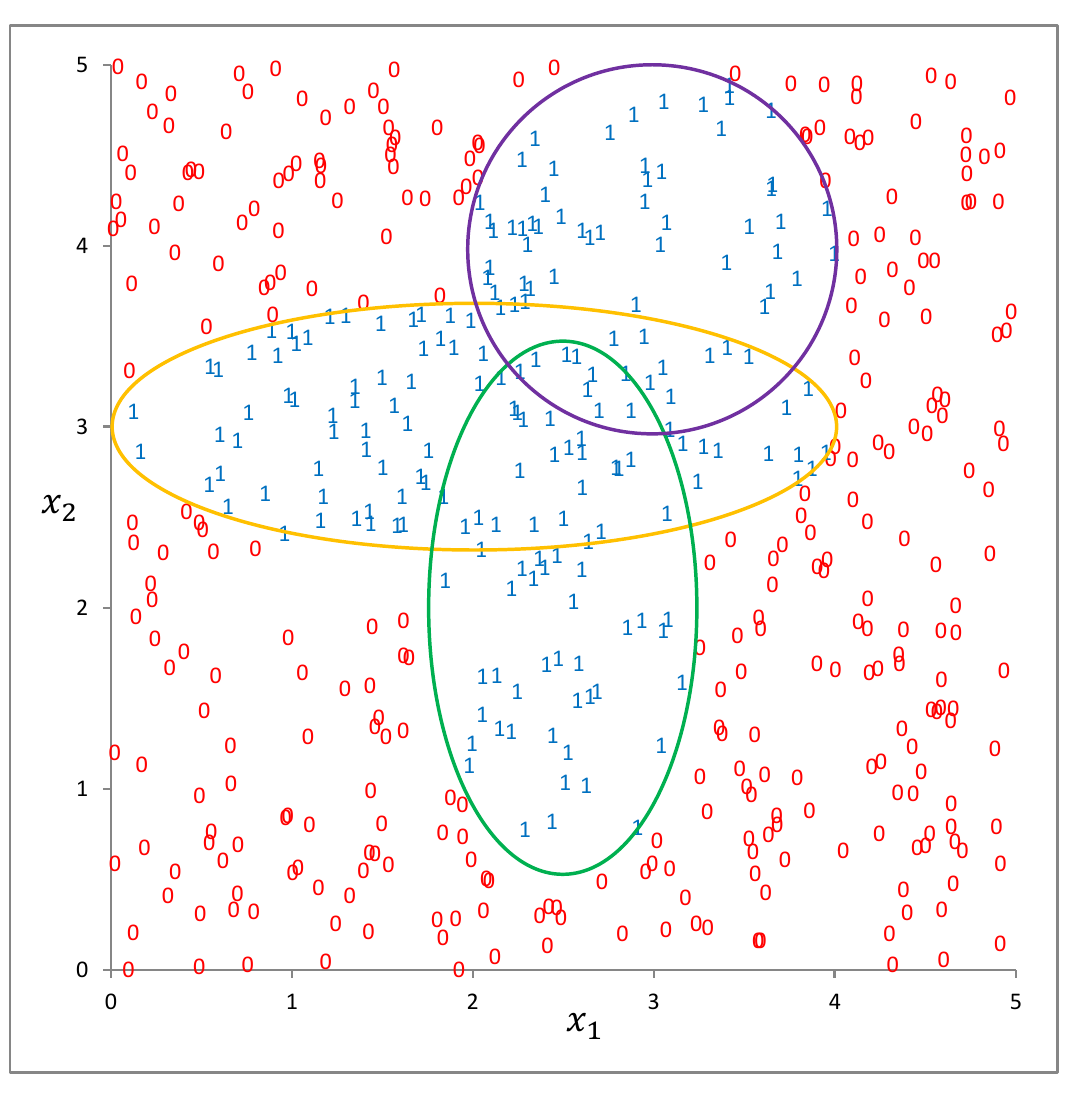}
\caption{Decision boundaries formed by Eq. (\ref{fun_mat_elly}), Eq. (\ref{fun_mat_ellz}) and Eq. (\ref{func_act_exp1}).}
\label{fig:ellipse_2}
\end{figure}

A functional transfer neural network can represent decision boundaries constructed by ellipses and hyperbolas. Recall the mathematical definition of ellipses: A ellipse with a centre $ (c_1,c_2) $ and semi-axes $ a_1 $ and $ a_2 $ can be defined as $ \cfrac{(x_1-c_1)^2}{a_1^2} + \cfrac{(x_2-c_2)^2}{a_2^2} = 1 $. This equation can be rewritten to $ r \cdot (\cfrac{1}{a_1})^2 \cdot (x_1-c_1)^2 + r \cdot (\cfrac{1}{a_2})^2 \cdot (x_2-c_2)^2 + (-r) = 0 $, where $ r $ is a positive number. Let $ p_1^2 = r \cdot (\cfrac{1}{a_1})^2 $, $ p_2^2 = r \cdot (\cfrac{1}{a_2})^2 $, $ q_1 = c_1 $, $ q_2 = c_2 $ and $ b = -r $. The equation becomes $ p_1^2 \cdot (x_1-q_1)^2 + p_2^2 \cdot (x_2-q_2)^2 + b = 0 $. It is noticeable that $ p_1^2 \cdot (x_1-q_1)^2 + p_2^2 \cdot (x_2-q_2)^2 $ can be rewritten to $ \bm{(} ~p_1^2 \cdot (x-q_1)^2 ~~~ p_2^2 \cdot (x-q_2)^2~ \bm{)} \odot \begin{pmatrix} x_1 \\ x_2 \end{pmatrix} $, which is a transfer computation (See Eq. \ref{def_fmat_vec}). Therefore, a model with an activation $\xi(u)$ can be defined as $ y = \xi(\bm{(} ~p_1^2 \cdot (x-q_1)^2 ~~~ p_2^2 \cdot (x-q_2)^2~ \bm{)} \odot \begin{pmatrix} x_1 \\ x_2 \end{pmatrix} + b) $
where $ x_1 $ and $ x_2$ are inputs and $ y $ is an output. Based on the above structure, multiple ellipse decision boundaries can be formed. For instance:
\begin{equation}
\label{fun_mat_elly}
\begin{pmatrix}
y_1 \\
y_2 \\
y_3 \\
\end{pmatrix}=\xi(
\begin{pmatrix}
0.50^2\cdot(x-2.00)^2 ~&~ 1.41^2\cdot(x-3.00)^2 \\
1.33^2\cdot(x-2.50)^2 ~&~ 0.67^2\cdot(x-2.00)^2 \\
1.00^2\cdot(x-3.00)^2 ~&~ 1.00^2\cdot(x-4.00)^2 \\
\end{pmatrix}
\odot
\begin{pmatrix}
x_1 \\
x_2 \\
\end{pmatrix}
+
\begin{pmatrix}
-1.00 \\
-1.00 \\
-1.00 \\
\end{pmatrix}
)~~,
\end{equation}
\begin{equation}
\label{fun_mat_ellz}
z=\xi(
\begin{pmatrix}
-1.00 ~&~ -1.00 ~&~ -1.00 \\
\end{pmatrix}
\cdot
\begin{pmatrix}
y_1 \\
y_2 \\
y_3 \\
\end{pmatrix}
+
2.50
)~~,
\end{equation}
and
\begin{equation}
\label{func_act_exp1}
\xi(x)=
\begin{cases}
1, &x>=0\cr 0, &x<0
\end{cases}~~.
\end{equation}
In the above model, $ x_1 $ and $ x_2 $ are input nodes. $ y_1 $, $ y_2 $ and $ y_3 $ are hidden nodes. $ z $ is an output node. $ \xi(x) $ is an activation. The input nodes and the hidden nodes are connected via a functional transfer matrix. The hidden nodes and the output node are connected via a weight matrix. The hidden nodes and the output node are activated via biases and the activation. Fig. \ref{fig:ellipse_2} shows decision boundaries formed by this model: The functional transfer matrix and the hidden nodes form three ellipse boundaries (in orange, green and purple respectively). The weight matrix and the output node form a boundary which is the union of inner parts of all ellipse boundaries. In addition, this figure shows some example inputs and outputs: If the inputs are inside the decision boundaries, the output is 1. Otherwise, the output is 0.
The reason why the model can construct ellipse boundaries is that its functional transfer matrix consists of the following function:
\begin{equation}
\label{ell_funcmat}
F_{i,j}(x) = p_{i,j,1}^2 \cdot (x - p_{i,j,2})^2~~,
\end{equation}
where $ p_{i,j,1} $ and $ p_{i,j,2} $ are trainable parameters. If the input dimension of this matrix is 2, it will construct ellipse boundaries on a plane. If its input dimension is 3, it will construct ellipsoid boundaries in a 3-dimensional space. Generally, if its input dimension is $ N $ ($N >= 2$), it will construct boundaries represented by $ (N-1) $-dimensional closed hypersurfaces of ellipses in an $ N $-dimensional space. This phenomenon reflects a significant difference between the a transfer matrix with Eq. (\ref{ell_funcmat}) and a standard linear weight matrix, as the former models closed hypersurfaces, while the later models hyperplanes. In addition, to use the back-propagation algorithm to train the parameters, derivatives are computed via:
\begin{equation}
\label{ell_p1d}
\cfrac{\partial F_{i,j}(x)}{\partial p_{i,j,1}} = 2 \cdot p_{i,j,1} \cdot (x - p_{i,j,2})^2~~,
\end{equation}
\begin{equation}
\label{ell_p2d}
\cfrac{\partial F_{i,j}(x)}{\partial p_{i,j,2}} = -2 \cdot p_{i,j,1}^2 \cdot (x - p_{i,j,2})~~,
\end{equation}
\begin{equation}
\label{ell_fd}
\cfrac{\partial F_{i,j}(x)}{\partial x} = 2 \cdot p_{i,j,1}^2 \cdot (x - p_{i,j,2})~~,
\end{equation}

Further, an adapted version of Eq. (\ref{ell_funcmat}) can represent not only ellipses, but also hyperbolas. The adapted function is defined as:
\begin{equation}
\label{hb_funcmat}
F_{i,j}(x) = p_{i,j,1}^2 \cdot (x - p_{i,j,2})^2 \cdot u_{i,j}~~,
\end{equation}
where $ u_{i,j} $ is initialised as 1 or -1. Please note that $ u_{i,j} $ is a constant, but NOT a trainable parameter. Given a $ 1 \times 2 $ matrix with this function:
\begin{equation}
\bm{F_{conic}}=
\begin{pmatrix}
p_{1,1,1}^2 \cdot (x - p_{1,1,2})^2 \cdot u_{1,1}~~~p_{1,2,1}^2 \cdot (x - p_{1,2,2})^2 \cdot u_{1,2} \\
\end{pmatrix}~~.
\end{equation}
If it is initialised by:
\begin{equation}
\begin{pmatrix}
u_{1,1}~~~u_{1,2} \\
\end{pmatrix}=
\begin{pmatrix}
1~~~1 \\
\end{pmatrix}~~,
\end{equation}
then it can represent an ellipse boundary. On the other hand, if it is initialised by:
\begin{equation}
\begin{pmatrix}
u_{1,1}~~~u_{1,2} \\
\end{pmatrix}=
\begin{pmatrix}
1~-1 \\
\end{pmatrix}~~,
\end{equation}
then it can be used to form an hyperbola boundary. More generally, in an $ N $-dimensional space, an $ M \times N $ functional transfer matrix with Eq. (\ref{hb_funcmat}) can represent $ M $ different $ (N-1) $-dimensional conic hypersurfaces.

\subsection{Sleeping weights and dead weights}
\label{sec:sleep_dead_func}

In a standard neural network, connections between nodes usually do not ``sleep". The reason is that the connection from Node $ j $ to Node $ i $ is a real weight $ w_{i,j} $. Let $ x_{j} $ denote a state of Node $ j $. Node $ i $ will receive a signal $ w_{i,j} \cdot x_{j} $. When $ w_{i,j} \not = 0 $ and $ x_{j} \not = 0 $, the signal always influences Node $ i $, regardless of whether or not Node $ i $ needs it. To enable the connections to ``sleep" temporarily, the following function is used in a functional transfer matrix:
\begin{equation}
\label{relusgn_funcmat}
F_{i,j}(x) = rect(p_{i,j,1} \cdot x + p_{i,j,2}) \cdot u_{i,j}~~.
\end{equation}
In the above function, $ p_{i,j,1} $ and $ p_{i,j,2} $ are trainable parameters, and $ u_{i,j} $ is a constant which is initialised as 1 or -1 before training. It also makes use of the rectifier \cite{DBLP:journals/jmlr/GlorotBB11}:
\begin{equation}
\label{rectifier}
rect(x) =
\begin{cases}
x, &x \geq 0\cr 0, &x<0
\end{cases}~~.
\end{equation}
The rectifier makes the function be able to ``sleep". In other words, when $ p_{i,j,1} \cdot x + p_{i,j,2} \leq 0 $, $ F_{i,j}(x) $ must be zero. To use the back-propagation algorithm to train the parameters, derivatives are computed via:
\begin{equation}
\label{relu_p1d}
\cfrac{\partial F_{i,j}(x)}{\partial p_{i,j,1}} =
\begin{cases}
u_{i,j} \cdot x, & p_{i,j,1} \cdot x + p_{i,j,2} \geq 0
\cr 0, & p_{i,j,1} \cdot x + p_{i,j,2} < 0
\end{cases}~~,
\end{equation}
\begin{equation}
\label{relu_p2d}
\cfrac{\partial F_{i,j}(x)}{\partial p_{i,j,2}} = \begin{cases}
u_{i,j}, &p_{i,j,1} \cdot x + p_{i,j,2} \geq 0
\cr 0, &p_{i,j,1} \cdot x + p_{i,j,2} < 0
\end{cases}~~,
\end{equation}
\begin{equation}
\label{relu_fd}
\cfrac{\partial F_{i,j}(x)}{\partial x} = \begin{cases}
p_{i,j,1} \cdot u_{i,j}, &p_{i,j,1} \cdot x + p_{i,j,2} \geq 0
\cr 0, &p_{i,j,1} \cdot x + p_{i,j,2} <0
\end{cases}~~.
\end{equation}

Another function enables a connection between two nodes to ``die". The word ``die" is different from the word ``sleep", as the former means that $ F_{i,j}(x) = 0 $ for all $ x $, and the later means that $ F_{i,j}(x) = 0 $ for some $ x $. If the connection from Node $ j $ to Node $ i $ is dead, then any change of Node $ j $ does not influence Node $ i $. This function is:
\begin{equation}
\label{relusgnv2_funcmat}
F_{i,j}(x) = rect(p_{i,j}) \cdot u_{i,j} \cdot x~~.
\end{equation}
In the above function, $ p_{i,j} $ is a trainable parameter, and $ u_{i,j} $ is a constant which is initialised as 1 or -1 before training. To use the back-propagation algorithm to train the parameters, derivatives are computed via:
\begin{equation}
\label{relusgnv2_pd}
\cfrac{\partial F_{i,j}(x)}{\partial p_{i,j}} = \begin{cases}
u_{i,j} \cdot x, &p_{i,j} \geq 0
\cr 0, &p_{i,j} <0
\end{cases}~~,
\end{equation}
\begin{equation}
\label{relusgnv2_fd}
\cfrac{\partial F_{i,j}(x)}{\partial x} = rect(p_{i,j}) \cdot u_{i,j}~~.
\end{equation}
It is noticeable that $ F_{i,j}(x) $, $ \cfrac{\partial F_{i,j}(x)}{\partial p_{i,j}} $ and $ \cfrac{\partial F_{i,j}(x)}{\partial x} $ are zero when $ p_{i,j} < 0 $, which means that the function does not transfer any signal and cannot be updated in this case. In other words, the function is dead once $  p_{i,j} $ is updated to a negative value. A matrix with this function can then represent a partially connected neural network, as dead functions can be considered as broken connections after training. A problem about the function is that it is not able to ``revive". In other words, once $ p_{i,j} $ is updated to a negative value, it has no chance to be non-negative again. To solve this problem, an adapted version of the rectifier is used:
\begin{equation}
\label{rectifier_ada}
rect_{ada}(x) =
\begin{cases}
x, &x \geq 0\cr 0, &x<0
\end{cases}~~.
\end{equation}
For the computation of $ \cfrac{\partial F_{i,j}(x)}{\partial x} $, its derivative is:
\begin{equation}
\label{rectifier_dada1}
\cfrac{\partial rect_{ada}(x)}{\partial x} =
\begin{cases}
1, &x \geq 0\cr 0, &x<0
\end{cases}~~.
\end{equation}
For the computation of $ \cfrac{\partial F_{i,j}(x)}{\partial p_{i,j}} $, however, its derivative is arbitrarily defined as:
\begin{equation}
\label{rectifier_dada1}
\cfrac{\partial rect_{ada}(x)}{\partial x} \equiv 1~~.
\end{equation}
We use ``$ \equiv $" instead of ``$ = $" because this derivative is not a mathematically sound result, but a predefined value. Thus, derivatives of $ F_{i,j}(x) $ are computed via:
\begin{equation}
\label{adarelusgnv2_pd}
\cfrac{\partial F_{i,j}(x)}{\partial p_{i,j}} = u_{i,j} \cdot x~~,
\end{equation}
\begin{equation}
\label{adarelusgnv2_fd}
\cfrac{\partial F_{i,j}(x)}{\partial x} =
\begin{cases}
u_{i,j}, &p_{i,j} \geq 0
\cr 0, &p_{i,j} <0
\end{cases}~~.
\end{equation}
It is noticeable that $ \cfrac{\partial F_{i,j}(x)}{\partial p_{i,j}} $ is not zero when $ x $ is not zero, which means that $ p_{i,j} $ can be updated when it is negative. Thus, although the function is dead when $ p_{i,j} $ is negative, it can be revived by updating $ p_{i,j} $ to a non-negative value.

\subsection{Sequential Modelling via Memory Functions}
\label{sec:memory_func}

A functional transfer matrix with memory functions can be used to model sequences. For instance, given a sequential input $ \bm{x} = \{x_1,x_2,\cdots,x_T\} $, the $t$th state ($ t = 1,2,\cdots,T $) of a memory function $ F_{i,j}(x_t) $ is computed via:
\begin{equation}
\label{memory_funcmat}
F_{i,j}(x_t) = tanh(p_{i,j,1} \cdot x_t + p_{i,j,2} \cdot C_{i,j,t-1} + p_{i,j,3})~~,
\end{equation}
\begin{equation}
\label{memory_cell}
C_{i,j,t} = F_{i,j}(x_{t})~~,
\end{equation}
where $ p_{i,j,1} $, $ p_{i,j,2} $ and $ p_{i,j,3} $ are trainable parameters. In particular, $ C_{i,j,t} $ is a memory cell and its initial state $ C_{i,j,0} $ is set to 0. To use the back-propagation algorithm to train the parameters, derivatives are computed via:
\begin{equation}
\label{memory_funcmat_dp1}
\cfrac{\partial F_{i,j}(x_t)}{\partial p_{i,j,1}} = (1 - tanh(p_{i,j,1} \cdot x_t + p_{i,j,2} \cdot C_{i,j,t-1} + p_{i,j,3})^2) \cdot x~~,
\end{equation}
\begin{equation}
\label{memory_funcmat_dp2}
\cfrac{\partial F_{i,j}(x_t)}{\partial p_{i,j,2}} = (1 - tanh(p_{i,j,1} \cdot x_t + p_{i,j,2} \cdot C_{i,j,t-1} + p_{i,j,3})^2) \cdot C_{i,j,t-1}~~,
\end{equation}
\begin{equation}
\label{memory_funcmat_dp3}
\cfrac{\partial F_{i,j}(x_t)}{\partial p_{i,j,3}} = 1 - tanh(p_{i,j,1} \cdot x_t + p_{i,j,2} \cdot C_{i,j,t-1} + p_{i,j,3})^2~~,
\end{equation}
\begin{equation}
\label{memory_funcmat_df}
\cfrac{\partial F_{i,j}(x_t)}{\partial x_t} = (1 - tanh(p_{i,j,1} \cdot x_t + p_{i,j,2} \cdot C_{i,j,t-1} + p_{i,j,3})^2) \cdot p_{i,j,1}~~.
\end{equation}
An $ N \times M $ dimensional matrix with the memory function can record $ N \times M $ signals from a previous state, as each memory function records one signal. On the other hand, a standard recurrent neural network usually uses hidden units to record signals from the previous state. If it has $ N $ input units and $ M $ hidden units, then it can record $ M $ signals. It is noticeable that the functional transfer matrix with the memory function records $ N-1$ times more signals than the recurrent neural network.

\section{Practical training methods for functional transfer neural networks}
\label{sec:training}
This section discusses some practical training methods for functional transfer neural networks. These methods are also used in the experiments (Section \ref{sec:exp}).

\subsection{The model structure and initialisation}
\label{sec:model_structure}

\begin{figure}
\centering
\includegraphics[width=10cm]{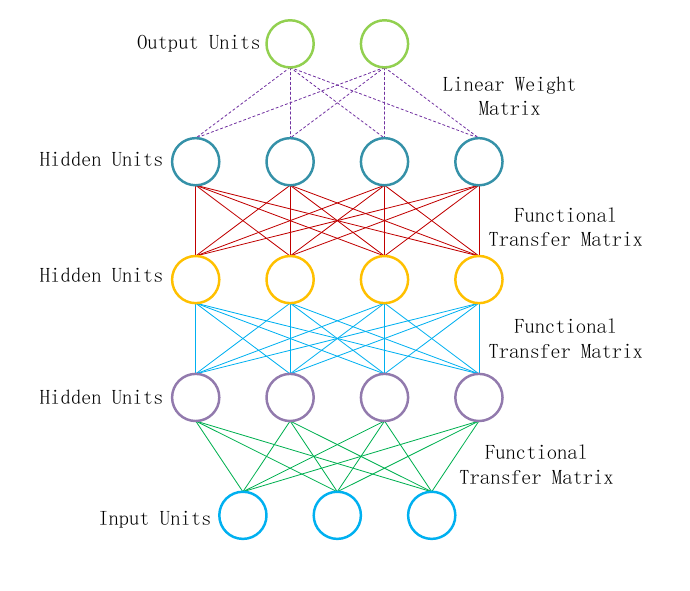}
\caption{The general structure of functional transfer neural networks.}
\label{fig:ftnn}
\end{figure}

Fig. \ref{fig:ftnn} shows the general structure of functional transfer neural networks: They usually have an input layer, one or more hidden layers and an output layer. Each hidden layer consists of a functional transfer matrix and some hidden units with a bias vector and an activation function. In particular, the activation function can be the logistic sigmoid function \cite{DBLP:journals/jmlr/GlorotBB11}:
\begin{equation}
logistic(u) = \cfrac{1}{1+e^{-u}}~~,
\end{equation}
the rectified linear unit (ReLU)
\begin{equation}
relu(u) =
\begin{cases}
u, &u \geq 0\cr 0, &u<0
\end{cases}~~,
\end{equation}
or the hyperbolic tangent (tanh) function
\begin{equation}
tanh(u) = \cfrac{e^{u}-e^{-u}}{e^{u}+e^{-u}}~~.
\end{equation}
The output layer consists of a linear weight matrix and $ N $ output units with a bias vector and a softmax function:
\begin{equation}
softmax(u_i) = \cfrac{e^{u_i}}{\sum \limits_{k=1}^N e^{u_k}}~~.
\end{equation}

After setting up a model based on the above structure, parameters are initialised based on the following method:
\begin{itemize}
\item Weights in the linear weight matrix are randomised as small real numbers.
\item Biases in the bias vectors are set to zero.
\item For different functional transfer matrices, different initialisation methods should be used. In the experimental section, some examples of initialisation are provided (see Table \ref{tab:exp_func} and Table \ref{tab:exp_init}).
\end{itemize}

\subsection{Training a single hidden layer}
\label{sec:train_single}
If a model only has one hidden layer, the training can be done via back-propagation directly \cite{rumelhart1988learning}: Firstly, an input signal is propagated to the output layer, and an output signal is generated. Then an error signal is computed by comparing the output signal with a target signal. The comparison is based on the cross-entropy criterion \cite{hinton2012deep}. Next, the error signal is back-propagated through the output layer and the hidden layer, and deltas of parameters are computed. For the output layer, the standard delta rules are used. For the hidden layer, the rules described by Section \ref{sec:backp} are used. Finally, the parameters are updated according to their deltas. The above process can be combined with stochastic gradient descent \cite{DBLP:conf/nips/ZinkevichWSL10}.
\subsection{Layer-wise supervised training and fine-tuning}
\label{sec:train_lwt_ft}

\begin{figure}
\centering
\includegraphics[width=12cm]{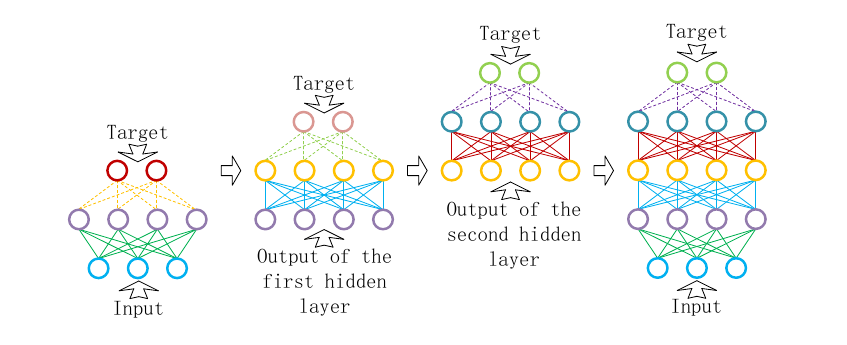}
\caption{Layer-wise supervised training and fine-tuning. Functional connections are denoted by solid lines, and linear weight connections are denoted by dashed lines.}
\label{fig:lwt}
\end{figure}

Training a functional transfer neural network can be more difficult when the number of hidden layers increases. To resolve this problem, the combination of layer-wise supervised training\footnote{For details about layer-wise supervised training, please also refer to Algorithm 7 in the appendix of the referenced paper.} and fine-tuning is used \cite{DBLP:conf/nips/BengioLPL06}, which is shown by Fig. \ref{fig:lwt}. Layer-wise training includes the following steps: Firstly, the first hidden layer is trained, while the other hidden layers and the output layer are ignored. To train this layer, a new softmax layer (with a linear weight matrix and a bias vector) is added onto it, and the method for training a single hidden layer (described by Section \ref{sec:train_single}) is used to train it. Then the softmax layer is removed, and the second hidden layer is added onto the first hidden layer. To train the second hidden layer, another new softmax layer is added onto it, and the same training method is used again. Please note that the first hidden layer is not updated in this step. Finally, all remaining hidden layers are trained by using the same method. In particular, the trained softmax layer on the last hidden layer is considered as the output layer of the whole neural network. After layer-wise training, the whole neural network can be further optimised via fine-tuning: For the output layer, the standard back-propagation rules are used; For the hidden layers, the rules described by Section \ref{sec:backp} are used. In practice, learning rates for fine-tuning are smaller than those for layer-wise training.

\section{Experiments}
\label{sec:exp}

\subsection{MNIST handwritten digit recognition}
\label{sec:mnistexp}

\subsubsection{Experimental settings}

\begin{table}
\centering
\caption{Example Functions Used to Model Functional Networks.}
\begin{tabular}{lll}
\hline\noalign{\smallskip}
ID & $ F_{i,j}(x) $ & Trainable Parameter(s) \\
\noalign{\smallskip}\hline\noalign{\smallskip}
F01 & $ p_{i,j}^2 \cdot x $ & $ p_{i,j} $ \\
F02 & $ p_{i,j}^3 \cdot x $ & $ p_{i,j} $ \\
F03 & $ p_{i,j} \cdot x^2 + q_{i,j} \cdot x $ & $ p_{i,j},q_{i,j} $ \\
F04 & $ p_{i,j} \cdot x^3 + q_{i,j} \cdot x^2 + r_{i,j} \cdot x $ & $ p_{i,j},q_{i,j},r_{i,j} $ \\
F05 & $ p_{i,j} \cdot e^{q_{i,j} \cdot x} $ & $ p_{i,j},q_{i,j} $ \\
F06 & $ rect(p_{i,j}) \cdot u_{i,j} \cdot x $ & $ p_{i,j} $ \\
F07 & $ rect_{ada}(p_{i,j}) \cdot u_{i,j} \cdot x $ & $ p_{i,j} $ \\
F08 & $ rect(p_{i,j} \cdot x+q_{i,j}) \cdot u_{i,j} $ & $ p_{i,j},q_{i,j} $ \\
F09 & $ p_{i,j} \cdot rect(q_{i,j} \cdot x+r_{i,j}) $ & $ p_{i,j},q_{i,j},r_{i,j} $ \\
F10 & $  logistic(p_{i,j} \cdot x+q_{i,j}) \cdot u_{i,j} $ & $ p_{i,j},q_{i,j} $ \\
F11 & $ p_{i,j} \cdot logistic(q_{i,j} \cdot x+r_{i,j}) $ & $ p_{i,j},q_{i,j},r_{i,j} $ \\
F12 & $ p_{i,j} \cdot sin(q_{i,j} \cdot x+r_{i,j}) $ & $ p_{i,j},q_{i,j},r_{i,j} $ \\
F13 & $ p_{i,j} \cdot cos(q_{i,j} \cdot x+r_{i,j}) $ & $ p_{i,j},q_{i,j},r_{i,j} $ \\
F14 & $ p_{i,j} \cdot sinh(q_{i,j} \cdot x+r_{i,j}) $ & $ p_{i,j},q_{i,j},r_{i,j} $ \\
F15 & $ p_{i,j} \cdot (cosh(q_{i,j} \cdot x+r_{i,j}) - 1) $ & $ p_{i,j},q_{i,j},r_{i,j} $ \\
F16 & $ p_{i,j} \cdot tanh(q_{i,j} \cdot x+r_{i,j}) $ & $ p_{i,j},q_{i,j},r_{i,j} $ \\
F17 & $ p_{i,j}^2 \cdot x^2 $ & $ p_{i,j} $ \\
F18 & $ p_{i,j}^2 \cdot x^2 \cdot u_{i,j} $ & $ p_{i,j} $ \\
F19 & $ p_{i,j}^2 \cdot (x-q_{i,j})^2 $ & $ p_{i,j},q_{i,j} $ \\
F20 & $ p_{i,j}^2 \cdot (x-q_{i,j})^2 \cdot u_{i,j} $ & $ p_{i,j},q_{i,j} $ \\
\noalign{\smallskip}\hline
\end{tabular}
\label{tab:exp_func}
\end{table}

Table \ref{tab:exp_func} provides 20 example functions which are used to construct functional transfer neural networks. $ F_{i,j}(x) $ in the functional matrix (See Eq. (\ref{eq_fun_mat})) is substituted for these example functions. In particular, $ p_{i,j} $, $ q_{i,j} $ and $ r_{i,j} $ are used to denote trainable parameters instead of using $ p_{i,j,k} $. These functions include all functions discussed in Section \ref{sec:func_exam}, except the memory function in Section \ref{sec:memory_func}: F06, F07, and F08 have been discussed in Section \ref{sec:sleep_dead_func}; F12 has been discussed in Section \ref{sec:per_func}; F19 and F20 have been discussed in Section \ref{sec:conic_func}. They also include functions which have not been discussed before, such as F01 - F05, F09 - F11 and F13 - F18. We test these examples to explore the flexibility of the choice of functions. In fact, a wider range of functions may be useful, but it is difficult to cover all possible functions in this paper. On the other hand, some kinds of functions may not work. For instance, we have found that $ F_{i,j}(x) = p_{i,j} \cdot cosh(q_{i,j} \cdot x+r_{i,j}) $ does not work, because the range of $ f(x) = cosh(x) $ is $ [1,+\infty) $, which means that this function does not have a ``zero state". To make it work, an additional ``-1" should be used to change the range to $ [0,+\infty) $, as indicated by F15. For another instance, $ f(x)=tan(x) $ does not work, because it is a discontinuous function. This is the reason why F12 cannot be changed to $ F_{i,j}(x) = p_{i,j} \cdot tan(q_{i,j} \cdot x+r_{i,j}) $. For the same reason, some other discontinuous functions, including $ f(x)=cot(x) $, $ f(x)=sec(x) $, $ f(x)=csc(x) $, $ f(x)=coth(x) $, $ f(x)=sech(x) $ and $ f(x)=csch(x) $, cannot be used in functional networks.

The model structure and the initialisation method have been revealed by Section \ref{sec:model_structure}. In particular, the input layer has 784 nodes, and the output layer has 10 nodes. They correspond to input pixels and output labels of the MNIST database \cite{DBLP:journals/spm/Deng12}. Besides, a model has 1, 5 or 10 hidden layers, and each hidden layer has 128 hidden units with the logistic activation, the ReLU or the tanh activation. Parameters of functional transfer matrices in these hidden layers are initialised based on Table \ref{tab:exp_init}. All training processes make use of the layer-wise supervised training and fine-tuning strategy discussed in Section \ref{sec:train_lwt_ft}. In particular, for layer-wise supervised training, the number of epochs is set to 15, and the learning rate is set to a constant $ 2^\gamma $, where $ \gamma \in \{0,-1,-2,-3,-4,-5\} $. For fine-tuning, the Newbob+/Train strategy is used \cite{DBLP:conf/icassp/WieslerRSN14}: The learning rate is set to $ 2^{\gamma - 4} $ initially, and it will be halved if the improvement of the cross-entropy loss is smaller than $ 10^{-4} $. The training stops when the improvement is smaller than zero, or after the 15th fine-tuning epoch.
For both layer-wise training and fine-tuning, the size of mini-batches is set to 16.

\begin{table}
\centering
\caption{Initialisation of Transfer Matrices.}
\begin{tabular}{lll}
\hline\noalign{\smallskip}
ID & Initialisation Range \\
\noalign{\smallskip}\hline\noalign{\smallskip}
F05 & $ -0.1 \leq p_{i,j} \leq 0.1,-4.0 \leq q_{i,j} \leq -2.0 $ \\
F06 & $ 0 \leq p_{i,j} \leq 2.0,u_{i,j} \in \{-1,1\} $ \\
F07 & $ 0 \leq p_{i,j} \leq 2.0,u_{i,j} \in \{-1,1\} $ \\
F08 & $ -0.1 \leq p_{i,j} \leq 0.1,-0.1 \leq q_{i,j} \leq 0.1,u_{i,j} \in \{-1,1\} $ \\
F10 & $ -0.1 \leq p_{i,j} \leq 0.1,-0.1 \leq q_{i,j} \leq 0.1,u_{i,j} \in \{-1,1\} $ \\
F12 & $ -0.1 \leq p_{i,j} \leq 0.1,-10.0 \leq q_{i,j} \leq 10.0,-10.0 \leq r_{i,j} \leq 10.0 $ \\
F13 & $ -0.1 \leq p_{i,j} \leq 0.1,-10.0 \leq q_{i,j} \leq 10.0,-10.0 \leq r_{i,j} \leq 10.0 $ \\
F18 & $ -0.1 \leq p_{i,j} \leq 0.1,u_{i,j} \in \{-1,1\} $ \\
F20 & $ -0.1 \leq p_{i,j} \leq 0.1,-0.1 \leq q_{i,j} \leq 0.1,u_{i,j} \in \{-1,1\} $ \\
Others & $ -0.1 \leq para \leq 0.1 $, where $ para $ denotes $ p_{i,j} $, $ q_{i,j} $ or $ r_{i,j} $. \\
\noalign{\smallskip}\hline
\end{tabular}
\label{tab:exp_init}
\end{table}

\subsubsection{Results}

\begin{table}
\centering
\caption{Accuracies (\%) and the Best $ \gamma $ --- 1 Hidden Layer.}
\label{tab:mnist_1l}
\begin{tabular}{|c||c|c||c||c|c||c||c|c||c|}
\hline
\multirow{2}{*}{Function}
& \multicolumn{3}{|c||}{Logistic} & \multicolumn{3}{|c||}{ReLU} & \multicolumn{3}{|c|}{Tanh} \\
\cline{2-10}
& LWT & FT & $ \gamma $ & LWT & FT & $ \gamma $ & LWT & FT & $ \gamma $ \\
\hline
F01 & 95.63 & 96.60 & -1 & 96.30 & 96.73 & -2 & 96.63 & 97.29 & -1 \\
\hline
F02 & 95.64 & 96.44 & 0 & 96.28 & 96.70 & -1 & 96.67 & 96.81 & -1 \\
\hline
F03 & 97.98 & 98.04 & 0 & 98.04 & 98.05 & -2 & 97.70 & 97.86 & -4 \\
\hline
F04 & 97.69 & 97.77 & -2 & 97.98 & 97.98 & -3 & 97.61 & 97.65 & -4 \\
\hline
F05 & 96.41 & 97.51 & -3 & 96.49 & 97.63 & -5 & 96.41 & 97.19 & -5 \\
\hline
F06 & 97.82 & 97.86 & 0 & 97.71 & 97.84 & -3 & 97.64 & 97.76 & -3 \\
\hline
F07 & 97.99 & 97.99 & 0 & 98.05 & 98.12 & -3 & 97.79 & 97.79 & -2 \\
\hline
F08 & 96.47 & 97.26 & -2 & 95.77 & 97.22 & -4 & 94.97 & 97.17 & -4 \\
\hline
F09 & 95.79 & 96.98 & -1 & 96.38 & 97.46 & -1 & 96.72 & 97.21 & -2 \\
\hline
F10 & 93.76 & 96.04 & -1 & 94.81 & 96.65 & -4 & 94.22 & 95.57 & -4 \\
\hline
F11 & 61.06 & 72.77 & -4 & $ \times $ & $ \times $ & $ \times $ & 76.36 & 88.92 & -3 \\
\hline
F12 & 96.84 & 97.08 & -2 & 97.10 & 97.10 & -3 & 96.51 & 96.61 & -4 \\
\hline
F13 & 96.53 & 96.90 & -3 & 96.51 & 96.94 & -5 & 96.41 & 96.80 & -4 \\
\hline
F14 & 95.82 & 97.00 & 0 & 96.91 & 97.49 & -2 & 97.48 & 97.51 & -1 \\
\hline
F15 & 93.72 & 94.65 & -1 & 91.74 & 94.44 & -2 & 94.70 & 96.09 & -1 \\
\hline
F16 & 96.29 & 96.92 & -1 & 96.04 & 97.56 & -2 & 96.79 & 97.30 & -2 \\
\hline
F17 & 95.65 & 96.54 & -1 & 94.68 & 96.29 & -2 & 96.83 & 97.34 & -2 \\
\hline
F18 & 97.16 & 97.47 & 0 & 97.60 & 97.60 & -1 & 97.02 & 97.16 & -2 \\
\hline
F19 & 94.68 & 96.84 & -1 & 96.45 & 96.51 & -1 & 97.01 & 97.43 & -2 \\
\hline
F20 & 97.67 & 97.88 & 0 & 97.87 & 97.90 & -1 & 97.04 & 97.47 & -2 \\
\hline
\end{tabular}
\end{table}

Table \ref{tab:mnist_1l} shows test results of the models with 1 hidden layer. The results include accuracies after layer-wise training (LWT), accuracies after fine-tuning (FT) and the best $ \gamma $ for each model\footnote{For full results of all $ \gamma $ values, please refer to Table \ref{tab:app_1layer} in Appendix.}. Failed training cases are marked as ``$ \times $". In addition, F01 - F20 are IDs of functional transfer matrices, and Logistic, ReLU and Tanh are activations of hidden units. It is noticeable that most models are trained successfully, except the F11-ReLU model. Although this model fails to be trained, its counterparts, including the F11-Logistic model and the F11-Tanh model, are trained successfully. Also, it is noticeable that the best $ \gamma $ is different for different functions and activations, which means that the initial learning rate should be carefully selected for different models.

\begin{table}
\centering
\caption{Accuracies (\%) and the Best $ \gamma $ --- 5 Hidden Layers.}
\label{tab:mnist_5l}
\begin{tabular}{|c||c|c||c||c|c||c||c|c||c|}
\hline
\multirow{2}{*}{Function}
& \multicolumn{3}{|c||}{Logistic} & \multicolumn{3}{|c||}{ReLU} & \multicolumn{3}{|c|}{Tanh} \\
\cline{2-10}
& LWT & FT & $ \gamma $ & LWT & FT & $ \gamma $ & LWT & FT & $ \gamma $ \\
\hline
F01 & 94.55 & 95.78 & -2 & 95.82 & 96.04 & -3 & 96.27 & 96.80 & -3 \\
\hline
F02 & 94.47 & 95.41 & 0 & 96.10 & 96.39 & -2 & 95.93 & 96.40 & -1 \\
\hline
F03 & 98.08 & 98.10 & -2 & 91.69 & 91.69 & -5 & 97.92 & 97.92 & -3 \\
\hline
F04 & 97.73 & 97.78 & -2 & 89.49 & 89.49 & -5 & 97.90 & 97.93 & -4 \\
\hline
F05 & 97.45 & 97.55 & -3 & 95.50 & 96.22 & -5 & $ \times $ & $ \times $ & $ \times $ \\
\hline
F06 & 97.79 & 97.85 & 0 & 97.70 & 97.70 & -3 & 97.91 & 97.91 & -4 \\
\hline
F07 & 97.92 & 97.92 & 0 & 97.58 & 97.58 & -4 & 97.75 & 97.79 & -3 \\
\hline
F08 & 97.41 & 97.56 & -2 & 97.30 & 97.37 & -4 & 97.40 & 97.46 & -4 \\
\hline
F09 & 95.75 & 96.28 & -1 & 97.39 & 97.54 & -2 & 96.92 & 96.98 & -2 \\
\hline
F10 & 93.66 & 95.44 & -1 & 96.60 & 96.89 & -3 & 95.76 & 95.94 & -3 \\
\hline
F11 & $ \times $ & $ \times $ & $ \times $ & $ \times $ & $ \times $ & $ \times $ & 86.25 & 87.94 & -3 \\
\hline
F12 & 96.83 & 96.83 & -2 & 93.62 & 93.62 & -4 & 96.13 & 96.13 & -4 \\
\hline
F13 & 96.44 & 96.44 & -2 & 94.03 & 94.03 & -3 & 95.91 & 95.91 & -4 \\
\hline
F14 & 96.68 & 96.97 & -1 & $ \times $ & $ \times $ & $ \times $ & 97.35 & 97.41 & -2 \\
\hline
F15 & 87.68 & 93.17 & -1 & 90.21 & 90.21 & -3 & 95.45 & 96.00 & -1 \\
\hline
F16 & 96.72 & 96.73 & -1 & 97.37 & 97.47 & -3 & 97.32 & 97.38 & -2 \\
\hline
F17 & 93.98 & 95.37 & -1 & $ \times $ & $ \times $ & $ \times $ & 84.23 & 95.79 & -3 \\
\hline
F18 & 96.60 & 96.81 & -1 & $ \times $ & $ \times $ & $ \times $ & 61.73 & 96.20 & -1 \\
\hline
F19 & 94.33 & 96.41 & -1 & $ \times $ & $ \times $ & $ \times $ & 95.60 & 96.52 & -2 \\
\hline
F20 & 97.24 & 97.25 & 0 & $ \times $ & $ \times $ & $ \times $ & 96.72 & 96.72 & -1 \\
\hline
\end{tabular}
\end{table}

Table \ref{tab:mnist_5l} shows test results of the models with 5 hidden layers\footnote{For full results, please refer to Table \ref{tab:app_5layer} in Appendix.}. There are 8 models failed to be trained, which means that training is becoming more difficult when the number of hidden layers increases. On the other hand, there are 52 models trained successfully, which demonstrates that the revised back-propagation rules for functional transfer matrices are able to work for deep models. Among these models, 14 models obtain a better accuracy (after fine-tuning) than their counterparts in Table \ref{tab:mnist_1l}, whereas 38 models obtain a worse accuracy. Therefore, no evidence can show that more hidden layers must be better than less hidden layers, and vice versa. In addition, for a certain function, the best $ \gamma $ for ReLU is usually smaller than that for Logistic, and the best $ \gamma $ for Tanh is always smaller than that for Logistic. A similar phenomenon about $ \gamma $ has also appeared in Table \ref{tab:mnist_1l}.

\begin{table}
\centering
\caption{Accuracies (\%) and the Best $ \gamma $ --- 10 Hidden Layers.}
\label{tab:mnist_10l}
\begin{tabular}{|c||c|c||c||c|c||c||c|c||c|}
\hline
\multirow{2}{*}{Function}
& \multicolumn{3}{|c||}{Logistic} & \multicolumn{3}{|c||}{ReLU} & \multicolumn{3}{|c|}{Tanh} \\
\cline{2-10}
& LWT & FT & $ \gamma $ & LWT & FT & $ \gamma $ & LWT & FT & $ \gamma $ \\
\hline
F01 & 93.48 & 95.08 & -2 & 95.81 & 95.91 & -3 & 95.76 & 96.28 & -3 \\
\hline
F02 & 93.14 & 94.92 & -1 & 96.07 & 96.65 & -2 & 95.63 & 96.24 & -2 \\
\hline
F03 & 97.85 & 97.85 & -2 & $ \times $ & $ \times $ & $ \times $ & 97.84 & 97.87 & -4 \\
\hline
F04 & 97.78 & 97.78 & -1 & $ \times $ & $ \times $ & $ \times $ & 97.82 & 97.82 & -2 \\
\hline
F05 & 97.11 & 97.52 & -3 & 95.15 & 95.92 & -5 & $ \times $ & $ \times $ & $ \times $ \\
\hline
F06 & 98.08 & 98.08 & -1 & 97.50 & 97.55 & -5 & 97.69 & 97.70 & -3 \\
\hline
F07 & 98.09 & 98.09 & -1 & 97.64 & 97.72 & -4 & 97.71 & 97.73 & -4 \\
\hline
F08 & 97.38 & 97.45 & -2 & 97.26 & 97.27 & -4 & 97.29 & 97.53 & -4 \\
\hline
F09 & 95.67 & 96.59 & -1 & 96.96 & 97.31 & -1 & 96.84 & 97.00 & -2 \\
\hline
F10 & 94.68 & 95.53 & -1 & 96.66 & 96.94 & -4 & 95.49 & 95.85 & -3 \\
\hline
F11 & $ \times $ & $ \times $ & $ \times $ & $ \times $ & $ \times $ & $ \times $ & 84.83 & 86.49 & -4 \\
\hline
F12 & 96.52 & 96.52 & -2 & 91.63 & 91.63 & -4 & 95.26 & 95.26 & -4 \\
\hline
F13 & 96.84 & 96.84 & -2 & 92.20 & 92.20 & -3 & 95.36 & 95.36 & -4 \\
\hline
F14 & 96.50 & 96.83 & 0 & $ \times $ & $ \times $ & $ \times $ & 97.09 & 97.37 & -2 \\
\hline
F15 & 83.40 & 90.20 & -2 & $ \times $ & $ \times $ & $ \times $ & 94.12 & 94.12 & -1 \\
\hline
F16 & 96.22 & 96.70 & -1 & 97.33 & 97.45 & -1 & 96.88 & 97.13 & -2 \\
\hline
F17 & 93.44 & 95.21 & -2 & $ \times $ & $ \times $ & $ \times $ & 71.84 & 95.60 & -3 \\
\hline
F18 & 96.40 & 96.84 & -1 & $ \times $ & $ \times $ & $ \times $ & 60.29 & 95.82 & -3 \\
\hline
F19 & 93.75 & 94.80 & -3 & $ \times $ & $ \times $ & $ \times $ & 95.44 & 95.44 & -1 \\
\hline
F20 & 97.11 & 97.41 & -1 & $ \times $ & $ \times $ & $ \times $ & 96.03 & 96.03 & -1 \\
\hline
\end{tabular}
\end{table}

Table \ref{tab:mnist_10l} shows test results of the models with 10 hidden layers\footnote{For full results, please refer to Table \ref{tab:app_10layer} in Appendix.}. There are 11 models failed to be trained and 49 models trained successfully, which means that the functional transfer matrices still can work when there are up to 10 hidden layers, but training has become more difficult. Among the models trained successfully, 12 models obtain a better accuracy (after fine-tuning) than their counterparts in Table \ref{tab:mnist_5l}, 36 models obtain a worse accuracy, and one model obtains the same accuracy. These results show again that increasing the number of hidden layers can (but not always) bring about a worse accuracy. In addition, a comparison among the $ \gamma $ values in Table \ref{tab:mnist_1l}, Table \ref{tab:mnist_5l} and Table \ref{tab:mnist_10l} reveals that the best $ \gamma $ values are influenced by the functional transfer matrices, the activations and the number of hidden layers, but it is difficult to find a way to predict them. Therefore, tuning of $ \gamma $ is still required in practice.

\subsection{Memorising the digits of $ \pi $}

\subsubsection{Experimental settings}

This is a short experiment to evaluate the memory function discussed by Section \ref{sec:memory_func}, because it is a time-dependent function and cannot be evaluated by using the same methodology as the previous experiments. In this part, we explore if functional transfer neural networks with the memory function are able to model the sequence of the circumference ratio\footnote{ The circumference ratio is chosen as experimental data because it is an irrational number, containing a non-cyclical sequence of digits. The dataset can be downloaded from \url{https://github.com/cchrewrite/Functional-Transfer-Neural-Networks/blob/master/circumference-ratio-data.txt}.} $ \pi = 3.141592653589793238\cdots $. Specifically, given the first $ n-1 $ digits, the neural networks are expected to output the $ n $th digit. Each neural network has an input layer, a hidden layer and an output layer: The input layer has 10 nodes which correspond to digits $ 0,1,2,\cdots,9 $. The hidden layer has 128, 256, 512 or 1024 nodes, and it consists of a functional transfer matrix with the memory function, a bias vector and a logistic activation. The output layer has 10 nodes which correspond to digits $ 0,1,2,\cdots,9 $, and it consists of a linear weight matrix, a bias vector and a softmax activation. All trainable parameters in the two matrices are initialised as a real number in $ [-0.1,0.1] $. All biases are initialised as zero. The training method described by Section \ref{sec:train_single} is used to train the models. Each training epoch makes use of a sequence of training pairs $ <x_i,x_{i+1}>~(i=1,2,\cdots,D) $. For each training pair, $ x_i $ and $ x_{i+1} $ are one-hot encodings of the $ i $th and $ (i+1) $th digits of $ \pi $ respectively. $ x_i $ is an input, and $ x_{i+1} $ is a target of output. $ D $ is set to 200, 400 or 800. The number of training epochs is set to 5000. The learning rate is set to $2^{-4}$, and it is not changed during the whole training process. Model performance is evaluated by the accuracy of predicting $ x_{i+1} $.

\subsubsection{Results}

\begin{figure}
\centering
\includegraphics[width=8cm]{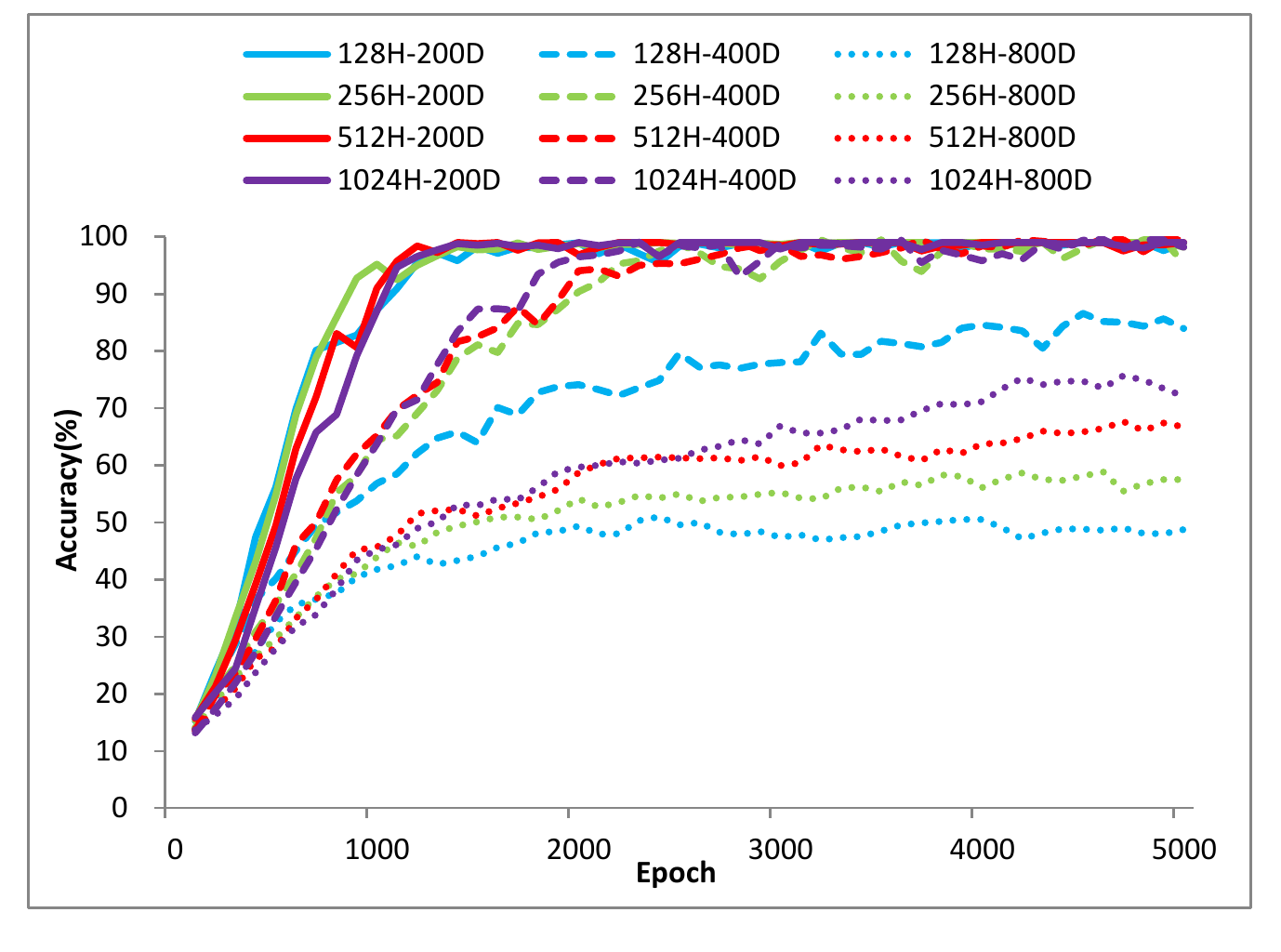}
\caption{Learning curves of functional transfer neural networks with the memory function.}
\label{fig:memory_res}
\end{figure}

Fig. \ref{fig:memory_res} shows accuracies of the models with the memory function. Each model is denoted by ``$x$H-$y$D", where $ x $ is the number of hidden units, and $ y $ the number of training pairs. It is noticeable that all 200D models gain high accuracies after the first 1500 training epochs, and the 256H-400D, 512H-400D and 1024H-400D models gain high accuracies after the first 2500 training epochs, which means that matrices with the memory function are able to model the sequence of digits. However, the 128H-400D model does not gain an accuracy as high as its counterparts, which means that its memory is restricted by the size of the matrix. In other words, to memorise a longer sequence, the number of hidden units should increase, because it determines the size of the functional transfer matrix and determines consequently the number of memory functions. For the 800D models, it is clear that the increase of the number of hidden units brings about better accuracies, though the accuracies are not significantly high.

\section{Related work}
\label{sec:relatedwork}

In this section, some previous work about functional-link neural networks, deep neural networks and neurons are revised, and comparisons between them and our work are carried out.

\subsection{Functional-link neural networks}
Functional-link neural networks use functional-links to enhance input patterns \cite{DBLP:journals/ijon/PaoPS94}. Usually, they consist of a functional expansion module, an input layer and an output layer, and the functional expansion module consists of many functional-links: For instance, a functional-link can be a trainable linear function (which is called a random variable functional-link), a multiplication function (which is called a generic basis), a trigonometric function or a Chebyshev polynomial basis function \cite{DBLP:journals/nca/DehuriC10}. A typical application of functional-link neural networks is to model nonlinear decision boundaries in channel equalisers \cite{DBLP:journals/sigpro/ZhaoZ08}. Dehuri and Cho \cite{DBLP:journals/eswa/DehuriC10} also use them as classifiers, where functional-links are used to select input features.

There are four main differences between our work and the above work about functional-link neural networks: Firstly, we use functional transfer matrices as alternatives to standard weight matrices, whereas functional expansion modules are NOT alternatives to weight matrices in functional-link neural networks. Secondly, our models have up to 10 hidden layers, and functional transfer matrices are applied to all of them, whereas functional-link neural networks have no hidden layer or have only one hidden layer with a linear weight matrix, and functional-links are only used to enhance input patterns \cite{DBLP:journals/eswa/DehuriC10,DBLP:journals/ijon/PaoPS94}. Thirdly, most functional transfer matrices are trainable, whereas most functional-links are fixed (except the random variable functional-link). Finally, functional transfer matrices are applied to hand-written digit recognition and the modelling of memory blocks, and they are different from the applications of functional-links.

\subsection{Deep architecture}
Deep architecture has been applied to many fields. For instance, it can be used to map pixels to a symbol \cite{DBLP:journals/jmlr/GlorotBB11,DBLP:journals/jmlr/GlorotB10}, map a frame of voice to a phone \cite{DBLP:journals/taslp/MohamedDH12,DBLP:conf/icassp/DengHK13}, model a sentence to a feature \cite{DBLP:journals/taslp/SarikayaHD14}, decide how to use a logical rule for automated theorem proving \cite{DBLP:conf/nips/IrvingSAECU16,DBLP:journals/corr/CaiKXS17}, manipulate variables in symbolic representations \cite{DBLP:conf/ijcnn/CaiKXS17}, guide the synthesis of programs \cite{DBLP:journals/corr/BalogGBNT16}, etc. In the above applications, deep neural networks play a role that maps input signals with different sizes, shapes and structures to vectorised features. Similarly, the first part of our experiments also focuses on this aspect: Neural networks with functional transfer matrices are used to map pixels to vectorised features.

\subsection{The improvement of neurons}
In a feedforward neural network, neurons (activations) in hidden layers are originally the logistic function \cite{lecun2015deep}. The logistic function can be replaced with the tanh function, the softsign function, the ReLU, the softplus function, the maxout function, or the soft-maxout function \cite{DBLP:journals/jmlr/GlorotBB11,bergstra2009quadratic,DBLP:conf/icassp/ZhangTPK14}. Also, neurons can be more complicated: For instance, they can be Clifford neurons in hyper-conformal space and be used to construct ellipsoidal hypersurfaces \cite{DBLP:conf/ijcnn/VillasenorAAL17}. The Clifford neurons are based on the Clifford algebra which extends multiplications of neural networks from real spaces to multi-dimensional geometric spaces \cite{DBLP:journals/nn/BuchholzS08}. Also, they can be the Taylor series of some simple activation functions, and the Taylor series can become trainable functions, because their coefficients can be trained \cite{DBLP:conf/ijcnn/ChungLP16}. Another kind of trainable neurons is called the Hermitian activation function \cite{DBLP:conf/icassp/SiniscalchiSSL10}. The Hermitian activation function is applied to feedforward neural networks with one hidden layer, and its shape can be changed by adapting its parameters \cite{DBLP:journals/taslp/SiniscalchiLL13}. Besides, neurons can be functions processing not only real numbers, but also complex numbers, according to the theory of complex back-propagation \cite{DBLP:journals/tsp/LeungH91a}: For instance, they can form multivalued neurons in feedforward neural networks \cite{DBLP:journals/soco/AizenbergM07}. They can also be extended to the hyper-complex domain and form quaternion neural networks, which can process input elements with a scaler and three orthogonal vectors \cite{DBLP:journals/jifs/MatsuiIKPN04}. Further, they can be applied to not only fully connected neural networks, but also convolutional neural networks \cite{DBLP:conf/ijcnn/KominamiOM17}. Different from the above work about neurons, our work focuses on improving the matrices between neurons instead of improving neurons themselves.

\section{Conclusion and future work}
\label{sec:con}

In this article, the theory and the applications of functional transfer matrices are presented. As argued in Section \ref{sec:fnn}, connections between nodes of a fully connected neural network can be represented by trainable functional connections instead of standard weights, and the theory of back-propagation can be extended to the training of these functional connections. Different functional transfer matrices can form different mathematical structures, such as conic hypersurfaces, periodic functions, sleeping units and dead units. Also, multiple functional transfer matrices can form a deep structure called a deep functional transfer neural network, but the training becomes more difficult when the number of hidden layers increases. Practically, the use of layer-wise supervised training and fine-tuning can solve this problem. In the experiments, a wide range of functional transfer matrices are demonstrated to be trainable and be able to gain high test accuracies on the MNIST database. It is also demonstrated that the neural network with the memory function can roughly memorise hundreds of digits of the circumference ratio, which means that the neural network can memorise previous information and form a memory block. There also remain many possibilities worth exploring in the future: Firstly, it is possible to apply functional transfer matrices to other kinds of neural networks, such as recurrent neural networks and convolutional neural networks, because their weights may also be replaced with functional connections. Secondly, it is often possible to design functional connections for special purposes, as there are an infinite number of possible combinations of functions. Thirdly, the training of deep functional transfer neural networks still relies on the layer-wise approach. Is there any efficient and effect way to train all layers directly? Finally, it is possible to apply functional transfer matrices to more fields besides the two applications in this work. In summary, this article has carried out the concept and shown the viability of changing the structure of neural networks by replacing weights with functional connections, and it is believed that the use of functional connections can bring about more potential for neural networks.

\section*{Acknowledgement}

Funding: This work was supported by the Fundamental Research Funds for the Central Universities [grant number 2016JX06]; and the National Natural Science Foundation of China [grant number 61472369].

\section*{References}

\bibliography{mybibfile}

\newpage

\section*{Appendix}
The appendix provides full results of the experiments in Section \ref{sec:mnistexp}: Table \ref{tab:app_1layer}, Table \ref{tab:app_5layer} and Table \ref{tab:app_10layer} shows test accuracies of models with 1, 5 and 10 hidden layers respectively. In these tables, accuracies are recorded as $ x/y $, where $ x $ denotes an accuracy after layer-wise supervised training, and $ y $ denotes an accuracy after fine-tuning. In particular, ``$ \times $" means that a model fails to be trained. To reduce the size of tables, ``Activation" is abbreviated to ``Act.", and the logistic sigmoid function is abbreviated to ``LogS".

\begin{table}[H]
\renewcommand\arraystretch{0.9}
\tiny
\centering
\caption{Full Results on MNIST --- 1 Hidden Layer}
\label{tab:app_1layer}
\begin{tabular}{|c|c|c|c|c|c|c|c|}
\hline
ID & Act. & $ \gamma = 0 $ & $ \gamma = -1 $ & $ \gamma = -2 $ & $ \gamma = -3 $ & $ \gamma = -4 $ & $ \gamma = -5 $ \\
\hline
\multirow{3}{*}{F01}
& LogS & 93.70/96.18 & 95.63/96.60 & 95.89/96.40 & 95.06/95.75 & 94.43/94.95 & 92.58/93.03 \\
\cline{2-8}
& ReLU & 94.26/94.51 & 96.51/96.66 & 96.30/96.73 & 94.96/95.77 & 93.33/93.91 & 92.06/92.23 \\
\cline{2-8}
& Tanh & 96.61/96.67 & 96.63/97.29 & 96.56/97.29 & 96.91/97.15 & 96.00/96.17 & 94.89/94.89 \\
\hline
\multirow{3}{*}{F02}
& LogS & 95.64/96.44 & 95.24/95.80 & 94.36/94.59 & 92.85/93.18 & 89.92/90.21 & 85.99/86.36 \\
\cline{2-8}
& ReLU & 74.19/75.23 & 96.28/96.70 & 95.20/95.98 & 94.76/95.43 & 93.25/93.55 & 90.69/91.21 \\
\cline{2-8}
& Tanh & 95.55/95.93 & 96.67/96.81 & 96.16/96.54 & 95.38/95.78 & 94.62/94.71 & 92.18/92.25 \\
\hline
\multirow{3}{*}{F03}
& LogS & 97.98/98.04 & 97.83/97.86 & 97.63/97.65 & 97.75/97.75 & 97.08/97.19 & 96.31/96.39 \\
\cline{2-8}
& ReLU & $ \times $ & $ \times $ & 98.04/98.05 & 97.91/97.92 & 97.98/97.98 & 97.70/97.88 \\
\cline{2-8}
& Tanh & 42.22/48.03 & 97.10/97.10 & 97.80/97.80 & 97.64/97.67 & 97.70/97.86 & 97.55/97.64 \\
\hline
\multirow{3}{*}{F04}
& LogS & 97.76/97.76 & 97.72/97.76 & 97.69/97.77 & 97.46/97.52 & 97.31/97.39 & 96.44/96.61 \\
\cline{2-8}
& ReLU & 46.72/47.80 & 96.68/96.78 & 94.68/94.68 & 97.98/97.98 & 97.78/97.89 & 97.59/97.74 \\
\cline{2-8}
& Tanh & 43.07/45.90 & 95.17/96.73 & 97.40/97.44 & 97.57/97.57 & 97.61/97.65 & 97.45/97.54 \\
\hline
\multirow{3}{*}{F05}
& LogS & 89.10/90.01 & 91.01/91.69 & 96.88/96.90 & 96.41/97.51 & 95.80/96.93 & 95.01/95.59 \\
\cline{2-8}
& ReLU & $ \times $ & $ \times $ & $ \times $ & $ \times $ & 95.61/97.29 & 96.49/97.63 \\
\cline{2-8}
& Tanh & $ \times $ & $ \times $ & $ \times $ & 79.79/90.40 & 96.15/96.21 & 96.41/97.19 \\
\hline
\multirow{3}{*}{F06}
& LogS & 97.82/97.86 & 97.74/97.82 & 97.69/97.69 & 97.35/97.43 & 96.43/96.60 & 95.45/95.53 \\
\cline{2-8}
& ReLU & 96.77/96.81 & 97.62/97.62 & 97.71/97.82 & 97.71/97.84 & 97.46/97.64 & 97.31/97.41 \\
\cline{2-8}
& Tanh & 41.00/41.86 & 97.37/97.43 & 97.61/97.67 & 97.64/97.76 & 97.41/97.57 & 97.12/97.40 \\
\hline
\multirow{3}{*}{F07}
& LogS & 97.99/97.99 & 97.76/97.89 & 97.74/97.74 & 97.34/97.43 & 96.69/96.75 & 95.36/95.44 \\
\cline{2-8}
& ReLU & 96.46/96.48 & 97.94/97.97 & 97.77/97.81 & 98.05/98.12 & 97.67/97.72 & 97.44/97.55 \\
\cline{2-8}
& Tanh & 45.75/49.80 & 96.91/97.43 & 97.79/97.79 & 97.53/97.61 & 97.34/97.47 & 97.25/97.28 \\
\hline
\multirow{3}{*}{F08}
& LogS & 93.55/93.90 & 95.50/95.66 & 96.47/97.26 & 95.65/96.93 & 94.89/96.12 & 93.82/94.46 \\
\cline{2-8}
& ReLU & $ \times $ & $ \times $ & $ \times $ & 68.57/76.95 & 95.77/97.22 & 95.52/96.98 \\
\cline{2-8}
& Tanh & $ \times $ & $ \times $ & $ \times $ & 94.54/94.69 & 94.97/97.17 & 95.86/96.76 \\
\hline
\multirow{3}{*}{F09}
& LogS & 95.03/96.85 & 95.79/96.98 & 95.42/96.05 & 93.98/94.53 & 92.81/93.02 & 90.64/90.83 \\
\cline{2-8}
& ReLU & 91.13/91.29 & 96.38/97.46 & 95.66/97.32 & 96.20/96.80 & 94.81/95.53 & 93.06/93.62 \\
\cline{2-8}
& Tanh & 96.55/96.61 & 95.80/97.07 & 96.72/97.21 & 95.73/96.49 & 94.49/95.11 & 93.66/93.79 \\
\hline
\multirow{3}{*}{F10}
& LogS & 93.40/93.69 & 93.76/96.04 & 93.09/95.63 & 93.59/94.99 & 91.64/92.80 & 91.14/91.91 \\
\cline{2-8}
& ReLU & $ \times $ & $ \times $ & $ \times $ & 93.62/95.45 & 94.81/96.65 & 94.78/95.61 \\
\cline{2-8}
& Tanh & $ \times $ & 55.44/61.14 & 93.47/93.58 & 92.33/95.28 & 94.22/95.57 & 91.85/92.97 \\
\hline
\multirow{3}{*}{F11}
& LogS & $ \times $ & $ \times $ & $ \times $ & $ \times $ & 61.06/72.77 & $ \times $ \\
\cline{2-8}
& ReLU & $ \times $ & $ \times $ & $ \times $ & $ \times $ & $ \times $ & $ \times $ \\
\cline{2-8}
& Tanh & $ \times $ & $ \times $ & $ \times $ & 76.36/88.92 & 79.61/88.04 & 78.72/86.42 \\
\hline
\multirow{3}{*}{F12}
& LogS & 93.05/93.40 & 96.75/96.75 & 96.84/97.08 & 96.62/96.82 & 95.85/96.21 & 94.99/95.29 \\
\cline{2-8}
& ReLU & $ \times $ & $ \times $ & 59.26/61.40 & 97.10/97.10 & 96.76/96.96 & 96.28/96.72 \\
\cline{2-8}
& Tanh & $ \times $ & $ \times $ & 88.46/90.36 & 96.12/96.22 & 96.51/96.61 & 96.02/96.54 \\
\hline
\multirow{3}{*}{F13}
& LogS & 92.81/93.33 & 96.22/96.29 & 96.55/96.80 & 96.53/96.90 & 96.14/96.53 & 95.24/95.55 \\
\cline{2-8}
& ReLU & $ \times $ & $ \times $ & 73.58/77.64 & 96.46/96.53 & 96.47/96.83 & 96.51/96.94 \\
\cline{2-8}
& Tanh & $ \times $ & $ \times $ & 87.93/91.68 & 95.58/95.61 & 96.41/96.80 & 95.92/96.22 \\
\hline
\multirow{3}{*}{F14}
& LogS & 95.82/97.00 & 95.87/96.86 & 95.81/96.55 & 94.22/94.98 & 93.41/93.61 & 91.12/91.62 \\
\cline{2-8}
& ReLU & $ \times $ & 96.82/97.45 & 96.91/97.49 & 96.50/97.20 & 95.55/96.24 & 94.25/94.84 \\
\cline{2-8}
& Tanh & 96.86/97.02 & 97.48/97.51 & 96.59/97.21 & 96.08/96.82 & 94.70/95.62 & 93.36/93.79 \\
\hline
\multirow{3}{*}{F15}
& LogS & 94.15/94.32 & 93.72/94.65 & 91.99/92.96 & 89.38/90.58 & $ \times $ & $ \times $ \\
\cline{2-8}
& ReLU & $ \times $ & 81.87/89.81 & 91.74/94.44 & 91.03/92.79 & 90.90/91.74 & 87.08/88.06 \\
\cline{2-8}
& Tanh & 94.38/94.60 & 94.70/96.09 & 93.69/95.04 & 93.63/94.34 & 91.79/91.92 & 88.86/89.02 \\
\hline
\multirow{3}{*}{F16}
& LogS & 95.44/96.76 & 96.29/96.92 & 95.96/96.40 & 94.86/95.11 & 93.18/93.56 & 90.81/91.22 \\
\cline{2-8}
& ReLU & $ \times $ & 96.19/97.54 & 96.04/97.56 & 96.26/97.07 & 95.33/95.92 & 94.25/94.75 \\
\cline{2-8}
& Tanh & 97.14/97.18 & 96.29/97.20 & 96.79/97.30 & 96.25/97.02 & 94.84/95.59 & 93.24/93.61 \\
\hline
\multirow{3}{*}{F17}
& LogS & 93.76/96.02 & 95.65/96.54 & 95.93/96.36 & 95.36/95.83 & 93.96/94.38 & 92.16/92.67 \\
\cline{2-8}
& ReLU & 95.21/95.32 & $ \times $ & 94.68/96.29 & 94.36/95.05 & 92.17/92.98 & 91.62/91.86 \\
\cline{2-8}
& Tanh & 96.79/96.83 & 96.35/97.24 & 96.83/97.34 & 96.89/96.97 & 95.87/95.97 & 94.72/94.98 \\
\hline
\multirow{3}{*}{F18}
& LogS & 97.16/97.47 & 97.11/97.37 & 96.65/96.89 & 95.60/95.78 & 94.24/94.49 & 92.06/92.22 \\
\cline{2-8}
& ReLU & 97.51/97.51 & 97.60/97.60 & 97.14/97.53 & 96.97/97.28 & 96.11/96.13 & 95.03/95.13 \\
\cline{2-8}
& Tanh & 96.71/96.73 & 96.91/97.09 & 97.02/97.16 & 96.30/96.56 & 96.20/96.23 & 94.65/94.74 \\
\hline
\multirow{3}{*}{F19}
& LogS & 96.14/96.22 & 94.68/96.84 & 95.76/96.50 & 95.52/96.27 & 94.25/94.72 & 92.72/93.04 \\
\cline{2-8}
& ReLU & $ \times $ & 96.45/96.51 & 94.55/96.27 & 94.03/94.99 & 92.54/92.98 & 91.58/92.05 \\
\cline{2-8}
& Tanh & 96.64/96.88 & 97.12/97.12 & 97.01/97.43 & 96.91/97.03 & 96.05/96.26 & 95.19/95.33 \\
\hline
\multirow{3}{*}{F20}
& LogS & 97.67/97.88 & 96.97/97.32 & 96.71/97.16 & 95.81/96.10 & 94.67/94.89 & 92.97/93.23 \\
\cline{2-8}
& ReLU & 83.72/89.56 & 97.87/97.90 & 96.76/97.65 & 96.69/97.33 & 96.57/96.70 & 94.71/94.84 \\
\cline{2-8}
& Tanh & $ \times $ & 97.21/97.28 & 97.04/97.47 & 96.94/97.22 & 96.28/96.44 & 95.00/95.04 \\
\hline
\end{tabular}
\end{table}

\begin{table}[H]
\renewcommand\arraystretch{0.9}
\tiny
\centering
\caption{Full Results on MNIST --- 5 Hidden Layers}
\label{tab:app_5layer}
\begin{tabular}{|c|c|c|c|c|c|c|c|}
\hline
ID & Act. & $ \gamma = 0 $ & $ \gamma = -1 $ & $ \gamma = -2 $ & $ \gamma = -3 $ & $ \gamma = -4 $ & $ \gamma = -5 $ \\
\hline
\multirow{3}{*}{F01}
& LogS & 92.42/94.82 & 93.77/95.55 & 94.55/95.78 & 93.04/94.42 & 90.54/91.96 & 86.72/88.39 \\
\cline{2-8}
& ReLU & $ \times $ & $ \times $ & $ \times $ & 95.82/96.04 & 92.83/94.14 & 91.14/92.58 \\
\cline{2-8}
& Tanh & $ \times $ & 95.47/95.72 & 96.04/96.61 & 96.27/96.80 & 95.26/95.77 & 93.92/94.45 \\
\hline
\multirow{3}{*}{F02}
& LogS & 94.47/95.41 & 93.42/95.10 & 91.74/93.86 & 90.25/92.14 & $ \times $ & $ \times $ \\
\cline{2-8}
& ReLU & $ \times $ & 79.09/90.11 & 96.10/96.39 & 94.15/95.48 & 92.05/93.67 & 89.13/91.50 \\
\cline{2-8}
& Tanh & 95.20/95.28 & 95.93/96.40 & 95.75/96.30 & 95.25/95.75 & 93.46/94.18 & 91.01/91.82 \\
\hline
\multirow{3}{*}{F03}
& LogS & 97.90/97.92 & 97.99/97.99 & 98.08/98.10 & 97.66/97.67 & 97.29/97.41 & 96.47/96.59 \\
\cline{2-8}
& ReLU & $ \times $ & $ \times $ & $ \times $ & $ \times $ & $ \times $ & 91.69/91.69 \\
\cline{2-8}
& Tanh & $ \times $ & $ \times $ & 97.78/97.78 & 97.92/97.92 & 97.84/97.85 & 97.79/97.82 \\
\hline
\multirow{3}{*}{F04}
& LogS & 97.69/97.71 & 97.72/97.72 & 97.73/97.78 & 97.61/97.72 & 97.60/97.64 & 96.83/96.99 \\
\cline{2-8}
& ReLU & $ \times $ & $ \times $ & $ \times $ & $ \times $ & $ \times $ & 89.49/89.49 \\
\cline{2-8}
& Tanh & $ \times $ & $ \times $ & 97.59/97.59 & 97.79/97.80 & 97.90/97.93 & 97.63/97.63 \\
\hline
\multirow{3}{*}{F05}
& LogS & 87.68/88.71 & 91.20/91.65 & 96.73/97.07 & 97.45/97.55 & 96.49/97.00 & 95.05/95.72 \\
\cline{2-8}
& ReLU & $ \times $ & $ \times $ & $ \times $ & $ \times $ & 95.20/95.93 & 95.50/96.22 \\
\cline{2-8}
& Tanh & $ \times $ & $ \times $ & $ \times $ & $ \times $ & $ \times $ & $ \times $ \\
\hline
\multirow{3}{*}{F06}
& LogS & 97.79/97.85 & 97.81/97.81 & 97.71/97.73 & 97.44/97.48 & 96.85/96.93 & 95.43/95.48 \\
\cline{2-8}
& ReLU & $ \times $ & $ \times $ & $ \times $ & 97.70/97.70 & $ \times $ & 97.30/97.39 \\
\cline{2-8}
& Tanh & 51.02/51.02 & 97.50/97.50 & 97.60/97.60 & 97.54/97.63 & 97.91/97.91 & 97.55/97.58 \\
\hline
\multirow{3}{*}{F07}
& LogS & 97.92/97.92 & 97.79/97.79 & 97.67/97.74 & 97.59/97.64 & 96.66/96.73 & 95.49/95.61 \\
\cline{2-8}
& ReLU & $ \times $ & $ \times $ & $ \times $ & $ \times $ & 97.58/97.58 & 97.31/97.33 \\
\cline{2-8}
& Tanh & $ \times $ & 97.63/97.63 & 97.66/97.66 & 97.75/97.79 & 97.52/97.54 & 97.41/97.41 \\
\hline
\multirow{3}{*}{F08}
& LogS & 93.90/94.24 & 95.61/95.88 & 97.41/97.56 & 96.96/97.27 & 96.03/96.36 & 94.17/94.58 \\
\cline{2-8}
& ReLU & $ \times $ & $ \times $ & $ \times $ & 92.35/92.97 & 97.30/97.37 & 97.30/97.35 \\
\cline{2-8}
& Tanh & $ \times $ & $ \times $ & $ \times $ & 95.53/96.13 & 97.40/97.46 & 97.29/97.34 \\
\hline
\multirow{3}{*}{F09}
& LogS & 95.97/96.24 & 95.75/96.28 & 94.95/95.58 & 93.75/94.34 & 90.07/91.51 & 80.88/86.51 \\
\cline{2-8}
& ReLU & 92.43/92.76 & 97.15/97.32 & 97.39/97.54 & 96.85/96.85 & 95.48/95.79 & 93.60/93.89 \\
\cline{2-8}
& Tanh & 96.17/96.52 & 96.45/96.73 & 96.92/96.98 & 96.53/96.63 & 95.24/95.64 & 92.85/93.39 \\
\hline
\multirow{3}{*}{F10}
& LogS & 91.86/93.49 & 93.66/95.44 & 93.00/94.78 & 90.78/93.13 & 84.66/88.50 & 78.85/83.61 \\
\cline{2-8}
& ReLU & $ \times $ & $ \times $ & $ \times $ & 96.60/96.89 & 96.64/96.82 & 95.13/95.67 \\
\cline{2-8}
& Tanh & $ \times $ & $ \times $ & 94.03/94.12 & 95.76/95.94 & 95.27/95.58 & 93.26/94.04 \\
\hline
\multirow{3}{*}{F11}
& LogS & $ \times $ & $ \times $ & $ \times $ & $ \times $ & $ \times $ & $ \times $ \\
\cline{2-8}
& ReLU & $ \times $ & $ \times $ & $ \times $ & $ \times $ & $ \times $ & $ \times $ \\
\cline{2-8}
& Tanh & $ \times $ & $ \times $ & $ \times $ & 86.25/87.94 & 84.39/86.63 & 82.31/85.15 \\
\hline
\multirow{3}{*}{F12}
& LogS & 91.56/91.56 & 96.04/96.04 & 96.83/96.83 & 96.31/96.33 & 96.55/96.55 & 95.92/95.98 \\
\cline{2-8}
& ReLU & $ \times $ & $ \times $ & $ \times $ & 93.51/93.51 & 93.62/93.62 & 93.04/93.04 \\
\cline{2-8}
& Tanh & $ \times $ & $ \times $ & 88.65/88.65 & 95.33/95.33 & 96.13/96.13 & 95.62/95.62 \\
\hline
\multirow{3}{*}{F13}
& LogS & 91.06/91.06 & 95.91/95.91 & 96.44/96.44 & 96.35/96.39 & 96.13/96.13 & 96.10/96.10 \\
\cline{2-8}
& ReLU & $ \times $ & $ \times $ & $ \times $ & 94.03/94.03 & 93.74/93.74 & 93.45/93.45 \\
\cline{2-8}
& Tanh & $ \times $ & $ \times $ & 87.31/87.31 & 95.10/95.10 & 95.91/95.91 & 95.65/95.65 \\
\hline
\multirow{3}{*}{F14}
& LogS & 96.62/96.85 & 96.68/96.97 & 96.16/96.32 & 94.39/94.87 & 91.31/92.49 & 88.04/89.99 \\
\cline{2-8}
& ReLU & $ \times $ & $ \times $ & $ \times $ & $ \times $ & $ \times $ & $ \times $ \\
\cline{2-8}
& Tanh & 96.31/96.91 & 97.20/97.25 & 97.35/97.41 & 96.89/96.97 & 95.96/96.01 & 93.97/94.15 \\
\hline
\multirow{3}{*}{F15}
& LogS & 87.41/92.13 & 87.68/93.17 & 84.54/90.67 & $ \times $ & $ \times $ & $ \times $ \\
\cline{2-8}
& ReLU & $ \times $ & $ \times $ & $ \times $ & 90.21/90.21 & 88.93/88.93 & $ \times $ \\
\cline{2-8}
& Tanh & 94.19/94.19 & 95.45/96.00 & 94.33/95.34 & 91.79/93.14 & 89.10/91.01 & 83.06/86.23 \\
\hline
\multirow{3}{*}{F16}
& LogS & 96.28/96.53 & 96.72/96.73 & 96.25/96.65 & 94.61/94.82 & 91.40/92.44 & 87.13/89.47 \\
\cline{2-8}
& ReLU & $ \times $ & 97.22/97.33 & 97.40/97.44 & 97.37/97.47 & 95.98/96.24 & 94.36/94.80 \\
\cline{2-8}
& Tanh & 96.33/96.51 & 96.84/97.01 & 97.32/97.38 & 96.46/96.68 & 95.64/95.70 & 93.18/93.82 \\
\hline
\multirow{3}{*}{F17}
& LogS & 91.95/94.24 & 93.98/95.37 & 94.55/95.21 & 93.48/94.77 & 91.67/92.94 & 87.56/88.93 \\
\cline{2-8}
& ReLU & $ \times $ & $ \times $ & $ \times $ & $ \times $ & $ \times $ & $ \times $ \\
\cline{2-8}
& Tanh & 63.88/93.22 & 67.87/95.10 & 78.94/95.57 & 84.23/95.79 & 85.49/94.42 & 81.06/93.23 \\
\hline
\multirow{3}{*}{F18}
& LogS & 96.66/96.73 & 96.60/96.81 & 96.30/96.47 & 95.08/95.44 & 92.93/93.78 & 89.06/91.47 \\
\cline{2-8}
& ReLU & $ \times $ & $ \times $ & $ \times $ & $ \times $ & $ \times $ & $ \times $ \\
\cline{2-8}
& Tanh & 61.54/95.91 & 61.73/96.20 & 73.97/95.78 & 81.92/95.73 & 84.87/94.35 & 82.47/93.99 \\
\hline
\multirow{3}{*}{F19}
& LogS & $ \times $ & 94.33/96.41 & 95.27/95.88 & 93.67/94.87 & 92.09/93.13 & 88.87/89.68 \\
\cline{2-8}
& ReLU & $ \times $ & $ \times $ & $ \times $ & $ \times $ & $ \times $ & $ \times $ \\
\cline{2-8}
& Tanh & $ \times $ & 96.02/96.02 & 95.60/96.52 & 95.91/96.15 & 94.99/95.91 & 94.13/94.67 \\
\hline
\multirow{3}{*}{F20}
& LogS & 97.24/97.25 & 97.02/97.04 & 96.68/96.85 & 95.74/95.97 & 94.00/94.59 & 91.13/92.20 \\
\cline{2-8}
& ReLU & $ \times $ & $ \times $ & $ \times $ & $ \times $ & $ \times $ & $ \times $ \\
\cline{2-8}
& Tanh & $ \times $ & 96.72/96.72 & 96.38/96.56 & 96.05/96.57 & 95.59/96.00 & 94.41/95.02 \\
\hline
\end{tabular}
\end{table}

\begin{table}[H]
\renewcommand\arraystretch{0.9}
\tiny
\centering
\caption{Full Results on MNIST --- 10 Hidden Layers}
\label{tab:app_10layer}
\begin{tabular}{|c|c|c|c|c|c|c|c|}
\hline
ID & Act. & $ \gamma = 0 $ & $ \gamma = -1 $ & $ \gamma = -2 $ & $ \gamma = -3 $ & $ \gamma = -4 $ & $ \gamma = -5 $ \\
\hline
\multirow{3}{*}{F01}
& LogS & 90.33/92.32 & 91.96/94.73 & 93.48/95.08 & 92.01/93.72 & 89.92/91.79 & 83.04/87.84 \\
\cline{2-8}
& ReLU & $ \times $ & $ \times $ & $ \times $ & 95.81/95.91 & 92.03/93.45 & 90.69/91.97 \\
\cline{2-8}
& Tanh & $ \times $ & 95.04/95.04 & 95.70/96.10 & 95.76/96.28 & 94.96/95.66 & 93.03/94.26 \\
\hline
\multirow{3}{*}{F02}
& LogS & 93.63/93.79 & 93.14/94.92 & 91.84/93.55 & 88.80/91.46 & $ \times $ & $ \times $ \\
\cline{2-8}
& ReLU & $ \times $ & $ \times $ & 96.07/96.65 & 95.14/95.74 & 92.92/93.68 & 91.06/92.05 \\
\cline{2-8}
& Tanh & 95.34/95.34 & 95.85/95.85 & 95.63/96.24 & 94.59/95.56 & 92.96/94.05 & 90.61/91.69 \\
\hline
\multirow{3}{*}{F03}
& LogS & 97.63/97.63 & 97.77/97.77 & 97.85/97.85 & 97.59/97.65 & 97.22/97.41 & 96.67/96.86 \\
\cline{2-8}
& ReLU & $ \times $ & $ \times $ & $ \times $ & $ \times $ & $ \times $ & $ \times $ \\
\cline{2-8}
& Tanh & $ \times $ & $ \times $ & 97.70/97.70 & 97.83/97.85 & 97.84/97.87 & 97.65/97.67 \\
\hline
\multirow{3}{*}{F04}
& LogS & 97.40/97.45 & 97.78/97.78 & 97.65/97.70 & 97.53/97.62 & 97.40/97.51 & 96.74/96.86 \\
\cline{2-8}
& ReLU & $ \times $ & $ \times $ & $ \times $ & $ \times $ & $ \times $ & $ \times $ \\
\cline{2-8}
& Tanh & $ \times $ & $ \times $ & 97.82/97.82 & 97.71/97.72 & 97.44/97.44 & 97.66/97.69 \\
\hline
\multirow{3}{*}{F05}
& LogS & 81.21/81.21 & 91.34/92.20 & 96.73/97.05 & 97.11/97.52 & 96.11/96.71 & 94.47/95.21 \\
\cline{2-8}
& ReLU & $ \times $ & $ \times $ & $ \times $ & $ \times $ & 93.04/93.04 & 95.15/95.92 \\
\cline{2-8}
& Tanh & $ \times $ & $ \times $ & $ \times $ & $ \times $ & $ \times $ & $ \times $ \\
\hline
\multirow{3}{*}{F06}
& LogS & 97.86/97.94 & 98.08/98.08 & 97.78/97.80 & 97.36/97.45 & 96.73/96.92 & 94.92/95.17 \\
\cline{2-8}
& ReLU & $ \times $ & $ \times $ & $ \times $ & $ \times $ & $ \times $ & 97.50/97.55 \\
\cline{2-8}
& Tanh & $ \times $ & 97.59/97.59 & 97.54/97.60 & 97.69/97.70 & 97.44/97.45 & 97.24/97.26 \\
\hline
\multirow{3}{*}{F07}
& LogS & 97.71/97.71 & 98.09/98.09 & 97.64/97.73 & 97.51/97.57 & 96.65/96.82 & 95.17/95.34 \\
\cline{2-8}
& ReLU & $ \times $ & $ \times $ & $ \times $ & $ \times $ & 97.64/97.72 & 97.50/97.50 \\
\cline{2-8}
& Tanh & $ \times $ & 97.30/97.31 & 97.55/97.62 & 97.63/97.63 & 97.71/97.73 & 97.39/97.40 \\
\hline
\multirow{3}{*}{F08}
& LogS & 93.68/94.10 & 96.11/96.16 & 97.38/97.45 & 96.88/97.07 & 95.82/96.05 & 93.68/94.31 \\
\cline{2-8}
& ReLU & $ \times $ & $ \times $ & $ \times $ & 73.04/77.96 & 97.26/97.27 & 97.27/97.27 \\
\cline{2-8}
& Tanh & $ \times $ & $ \times $ & $ \times $ & 95.84/96.19 & 97.29/97.53 & 97.41/97.53 \\
\hline
\multirow{3}{*}{F09}
& LogS & 95.83/96.11 & 95.67/96.59 & 94.87/95.52 & 93.21/94.16 & 90.31/91.91 & 82.49/87.59 \\
\cline{2-8}
& ReLU & 91.71/91.71 & 96.96/97.31 & 96.96/97.03 & 96.67/96.98 & 95.06/95.51 & 92.88/93.60 \\
\cline{2-8}
& Tanh & 95.47/95.47 & 96.47/96.85 & 96.84/97.00 & 95.87/96.27 & 95.10/95.60 & 92.81/93.52 \\
\hline
\multirow{3}{*}{F10}
& LogS & 91.58/92.36 & 94.68/95.53 & 91.84/94.42 & 90.79/93.02 & 79.73/84.32 & 62.24/70.82 \\
\cline{2-8}
& ReLU & $ \times $ & $ \times $ & $ \times $ & 94.97/95.25 & 96.66/96.94 & 94.86/95.27 \\
\cline{2-8}
& Tanh & $ \times $ & $ \times $ & 94.87/94.97 & 95.49/95.85 & 93.75/94.03 & 93.76/94.12 \\
\hline
\multirow{3}{*}{F11}
& LogS & $ \times $ & $ \times $ & $ \times $ & $ \times $ & $ \times $ & $ \times $ \\
\cline{2-8}
& ReLU & $ \times $ & $ \times $ & $ \times $ & $ \times $ & $ \times $ & $ \times $ \\
\cline{2-8}
& Tanh & $ \times $ & $ \times $ & $ \times $ & 83.86/83.86 & 84.83/86.49 & 81.61/84.68 \\
\hline
\multirow{3}{*}{F12}
& LogS & 91.84/91.84 & 96.18/96.18 & 96.52/96.52 & 96.49/96.49 & 96.32/96.32 & 95.64/95.64 \\
\cline{2-8}
& ReLU & $ \times $ & $ \times $ & $ \times $ & 91.39/91.39 & 91.63/91.63 & 90.70/90.70 \\
\cline{2-8}
& Tanh & $ \times $ & $ \times $ & 85.67/85.67 & 94.96/94.96 & 95.26/95.26 & 95.13/95.13 \\
\hline
\multirow{3}{*}{F13}
& LogS & 90.84/90.84 & 95.91/95.91 & 96.84/96.84 & 96.50/96.50 & 96.25/96.25 & 95.97/95.97 \\
\cline{2-8}
& ReLU & $ \times $ & $ \times $ & $ \times $ & 92.20/92.20 & 92.08/92.08 & 90.73/90.73 \\
\cline{2-8}
& Tanh & $ \times $ & $ \times $ & 86.73/86.73 & 94.47/94.47 & 95.36/95.36 & 95.22/95.22 \\
\hline
\multirow{3}{*}{F14}
& LogS & 96.50/96.83 & 96.42/96.73 & 95.48/95.96 & 94.46/94.76 & 90.75/92.57 & 86.01/89.92 \\
\cline{2-8}
& ReLU & $ \times $ & $ \times $ & $ \times $ & $ \times $ & $ \times $ & $ \times $ \\
\cline{2-8}
& Tanh & 96.09/96.09 & 96.74/97.05 & 97.09/97.37 & 96.84/97.05 & 95.26/95.57 & 93.32/93.95 \\
\hline
\multirow{3}{*}{F15}
& LogS & 88.76/88.76 & 86.05/86.05 & 83.40/90.20 & $ \times $ & $ \times $ & $ \times $ \\
\cline{2-8}
& ReLU & $ \times $ & $ \times $ & $ \times $ & $ \times $ & $ \times $ & $ \times $ \\
\cline{2-8}
& Tanh & 92.89/92.89 & 94.12/94.12 & 93.47/93.47 & 91.73/92.73 & 88.37/90.77 & 82.29/85.08 \\
\hline
\multirow{3}{*}{F16}
& LogS & 95.98/96.45 & 96.22/96.70 & 95.56/95.97 & 93.60/94.23 & 90.56/91.89 & 84.16/88.88 \\
\cline{2-8}
& ReLU & $ \times $ & 97.33/97.45 & 97.22/97.25 & 97.11/97.23 & 95.91/96.32 & 93.52/94.26 \\
\cline{2-8}
& Tanh & 96.39/96.39 & 96.49/96.68 & 96.88/97.13 & 96.43/96.68 & 95.08/95.47 & 93.25/93.77 \\
\hline
\multirow{3}{*}{F17}
& LogS & 90.20/90.20 & 93.24/94.46 & 93.44/95.21 & 92.86/94.12 & 89.86/91.93 & 85.66/86.88 \\
\cline{2-8}
& ReLU & $ \times $ & $ \times $ & $ \times $ & $ \times $ & $ \times $ & $ \times $ \\
\cline{2-8}
& Tanh & $ \times $ & $ \times $ & $ \times $ & 71.84/95.60 & 74.99/94.36 & 67.85/92.86 \\
\hline
\multirow{3}{*}{F18}
& LogS & 96.34/96.65 & 96.40/96.84 & 96.12/96.70 & 94.85/95.51 & 91.49/92.95 & 87.64/90.82 \\
\cline{2-8}
& ReLU & $ \times $ & $ \times $ & $ \times $ & $ \times $ & $ \times $ & $ \times $ \\
\cline{2-8}
& Tanh & $ \times $ & $ \times $ & 46.90/95.71 & 60.29/95.82 & 72.18/95.21 & 67.11/93.56 \\
\hline
\multirow{3}{*}{F19}
& LogS & $ \times $ & 92.27/93.91 & 93.20/94.08 & 93.75/94.80 & 90.26/92.94 & 87.45/89.11 \\
\cline{2-8}
& ReLU & $ \times $ & $ \times $ & $ \times $ & $ \times $ & $ \times $ & $ \times $ \\
\cline{2-8}
& Tanh & $ \times $ & 95.44/95.44 & 95.28/95.28 & 95.20/95.20 & 94.68/95.21 & 92.56/93.86 \\
\hline
\multirow{3}{*}{F20}
& LogS & 97.12/97.39 & 97.11/97.41 & 96.30/96.74 & 95.16/95.77 & 93.09/94.33 & 89.18/90.91 \\
\cline{2-8}
& ReLU & $ \times $ & $ \times $ & $ \times $ & $ \times $ & $ \times $ & $ \times $ \\
\cline{2-8}
& Tanh & $ \times $ & 96.03/96.03 & 95.75/95.75 & 95.96/95.96 & 95.03/95.43 & 92.59/93.79 \\
\hline
\end{tabular}
\end{table}

\end{document}